\documentclass{article}
\usepackage{iclr2027_conference,times}

\usepackage{amsmath,amsfonts,bm}

\def\plaineqref#1{\ref{#1}}
\newcommand{\E}{\mathbb{E}}
\newcommand{\R}{\mathbb{R}}

\usepackage{hyperref}
\usepackage{url}

\usepackage{booktabs}
\usepackage{amsfonts}
\usepackage{nicefrac}
\usepackage{microtype}
\usepackage[table]{xcolor}

\usepackage{makecell}

\usepackage{amsmath,amssymb}
\usepackage{graphicx}
\usepackage{adjustbox}
\usepackage{multirow}
\usepackage{subcaption}
\usepackage{siunitx}
\usepackage[ruled,vlined]{algorithm2e}

\DeclareMathOperator{\tr}{tr}
\newcommand{\N}{\mathcal{N}}
\newcommand{\Id}{\mathbf{I}}
\newcommand{\Laplacian}{\mathbf{L}}
\newcommand{\Ubf}{\mathbf{U}}
\newcommand{\Lambdabf}{\boldsymbol{\Lambda}}
\newcommand{\Sbf}{\boldsymbol{\Sigma}}

\title{Sensor Geometry as a Flow-Matching Prior\\for Multi-Channel Brain Signals}

\author{\textbf{Jaedong Hwang}\\
Massachusetts Institute of Technology\\
\texttt{jdhwang@mit.edu}
}

\iclrfinalcopy

\begin{document}

\maketitle

\begin{abstract}
Flow-matching models start from an isotropic Gaussian source, the standard choice when the correlation structure of the data is unknown in advance.
For multi-channel brain recordings, however, part of this structure is known in advance.
Electrodes sit at fixed positions on the head, and volume conduction through the skull and scalp makes nearby electrodes co-vary in a way that is shared across subjects.
Existing EEG generative models nonetheless leave the network to learn this from scratch.
We put this structure into the source instead.
From the sensor coordinates alone, we build a $k$-nearest-neighbor graph and take a graph-Mat\'ern function of its Laplacian as the source covariance, so the flow starts from spatially coherent patterns rather than channel-independent noise.
The change adds no learned parameters, works with any coupling and any drift network, and uses the same three hyperparameters on every dataset.
Across eight EEG datasets and four flow-matching methods, the graph-Mat\'ern source lowers the spectral discrepancy between generated and real signals in the five clinical bands (PSD-KL) on most datasets.
PSD-KL falls by $12\%$ to $17\%$ in geometric mean over datasets depending on the method and by up to $40\%$ on PhysioNet-MI, the densest montage.
We show that the improvement stems from the spatial eigenvectors of the local graph of sensor positions, since randomizing the eigenvectors while preserving the eigenvalue spectrum eliminates the gain.
Furthermore, a prior fitted directly to the empirical data covariance performs worse than isotropic noise.
The same construction applies unchanged to MEG, intracranial EEG with patient-specific grids, and a traffic-sensor network, lowering PSD-KL for every method on each\footnote{\url{https://jd730.github.io/projects/GraphPrior}}.

\end{abstract}

\section{Introduction}
\label{sec:intro}

Flow-matching generative models~\citep{lipman2023fm,albergo2023si,liu2023rectified,tong2024otcfm,tong2024sf2m} start from an isotropic Gaussian, $x_0\sim\N(0,\Id)$, and learn a drift that transports it to the data.
For natural images this is the standard choice, since the correlation structure of the data is not known in advance.
In multi-channel brain recordings, however, electrodes sit at fixed positions on the head, and volume conduction through skull and scalp makes nearby electrodes co-vary in a way that is shared across subjects~\citep{nunez2006electric}.
This structure is known before any data are seen, yet existing EEG generators, adversarial~\citep{hartmann2018eeggan} and diffusion-based~\citep{klein2024diffusionerp} alike, start from unstructured Gaussian noise and leave the sensor layout unused, so the network has to learn this structure from scratch.

\begin{figure}[t]
\centering
\includegraphics[width=\linewidth]{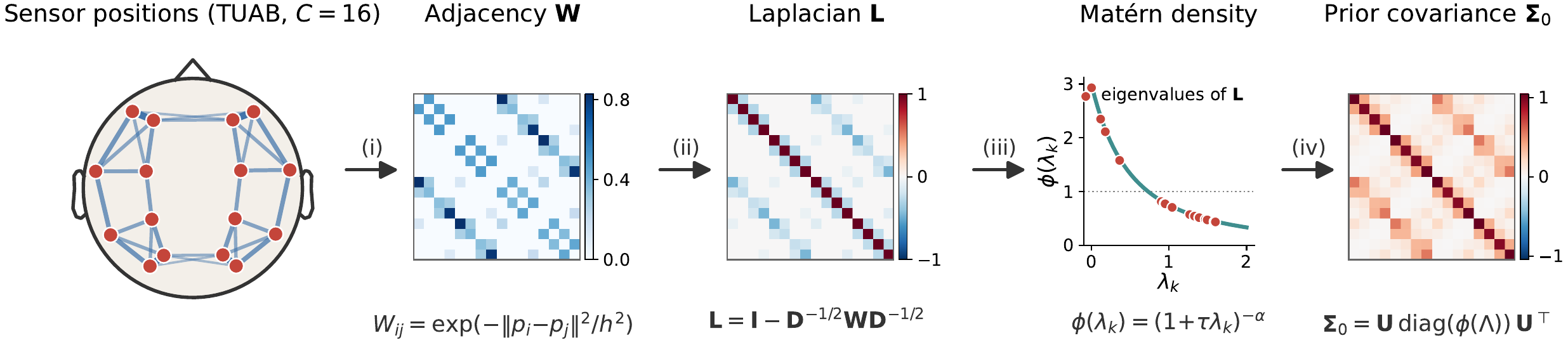}
\caption{\textbf{Building the graph-Mat\'ern source prior from sensor positions.}
(i)~Connect each sensor to its $k$ nearest neighbors, with edge weights that decay with distance, to get the adjacency matrix $\mathbf{W}$.
(ii)~Form the normalized graph Laplacian $\Laplacian$, whose eigenvectors are spatial patterns over the sensors ordered from smooth to rough.
(iii)~Weight each pattern by the Mat\'ern density $\phi(\lambda_m)=(1+\tau\lambda_m)^{-\alpha}$, which gives smooth patterns more variance and rough ones less.
(iv)~Assemble the resulting covariance $\Sbf_0$, from which the flow's source samples $x_0$ are drawn in Eq.~(\plaineqref{eq:prior}).
Every step uses only the sensor coordinates and never the recorded signals.}
\label{fig:prior-construction}
\end{figure}

We build this structure into the source instead, so that the flow starts from spatially coherent patterns and the drift does not have to learn the spatial correlations from scratch.
Concretely, we connect each sensor to its $k$ nearest neighbors, form the normalized Laplacian of that graph, and give each of its eigenvectors a variance set by a Mat\'ern density of its eigenvalue~\citep{borovitskiy2021graph} (see Figure~\ref{fig:prior-construction}).
The resulting graph-Mat\'ern prior puts most of its variance on spatial patterns that vary slowly across the scalp and little on patterns that flip sign between neighboring electrodes, so its samples are spatially coherent rather than channel-independent noise.
This modification affects only the source distribution.
It adds no learned parameters, works with any coupling and any drift network, and its three hyperparameters are fixed across all datasets.

We test the prior with four flow-matching methods, SF2M~\citep{tong2024sf2m}, SI~\citep{albergo2023si}, OT-CFM~\citep{tong2024otcfm}, and RF~\citep{liu2023rectified}, which cover the standard coupling choices, on eight EEG datasets, and measure how well generated signals reproduce the power spectrum of real signals in the five clinical frequency bands (PSD-KL; lower is better).
We also apply it to MEG, to intracranial EEG with patient-specific electrode grids~\citep{vanblooijs2023ccep}, and to the PEMS-BAY traffic-sensor network~\citep{li2018dcrnn}, to show that the prior is not specific to scalp EEG.
Replacing the prior's eigenvectors, spectrum, or graph while keeping the other two fixed shows that the gain in PSD-KL comes from the eigenvectors of the local graph of sensor positions.

Our contributions are as follows.
\begin{itemize}
\item We propose a source prior for flow matching computed from sensor coordinates alone---a graph-Mat\'ern covariance on the Laplacian of the graph of sensor positions---which replaces the standard isotropic source without adding learned parameters or modifying the drift network and coupling.
\item We show that the prior lowers the spectral discrepancy (PSD-KL) for four flow-matching methods on most of the eight EEG datasets, by up to $40\%$ on the densest montage, and that the same construction applies unchanged to MEG, intracranial EEG with patient-specific grids, and a traffic-sensor network.
\item On TUAB, we show that the improvement stems from the spatial eigenvectors of the local position graph, and that a prior fitted directly to the empirical data covariance performs worse than isotropic noise.
\end{itemize}


\section{Graph-Mat\'ern Source Prior for Flow Matching}
\label{sec:method}

\subsection{Preliminaries}
\label{subsec:prelim}

Let $p_0$ and $p_1$ be the prior and data distributions on $\R^{C\times T}$ (channels $\times$ time), and $\pi \in \Pi(p_0,p_1)$ be a coupling.
Flow matching \citep{lipman2023fm} trains a neural vector field $v_\theta(x,t)$ to match the conditional velocity $u_t(x_t\mid x_0,x_1) = \partial_t I(t,x_0,x_1)$ along an interpolation path parameterized by $t\sim\mathcal{U}(0,1)$; for the straight line $I(t,x_0,x_1)=(1-t)x_0+tx_1$ used in OT-CFM \citep{tong2024otcfm} this reduces to $u_t=x_1-x_0$, giving the squared-error loss of Eq.~(\plaineqref{eq:fm-loss}).
\begin{equation}
\mathcal{L}_{\mathrm{FM}} = \E_{t\sim\mathcal{U}(0,1),\,(x_0,x_1)\sim\pi,\, x_t=I(t,x_0,x_1)} \big\| v_\theta(x_t,t) - u_t \big\|_2^2.
\label{eq:fm-loss}
\end{equation}
Stochastic interpolants \citep{albergo2023si} add a latent Gaussian noise term, yielding $X_t = (1-t)x_0 + t x_1 + \gamma(t) z$ with $z\sim\N(0,\Id)$ and $\gamma(0)=\gamma(1)=0$ so that $X_0=x_0$, $X_1=x_1$. Differentiating in $t$ gives $u_t = (x_1-x_0) + \gamma'(t)z$.
The marginal drift $b(x,t)=\E[u_t\mid X_t=x]$ is the probability-flow ODE drift for the interpolant law \citep{albergo2023si}.
Finally, for a graph of sensor positions $G=(\mathcal{V},\mathcal{E})$ with $|\mathcal{V}|=C$, adjacency matrix $\mathbf{W}$, and degree matrix $\mathbf{D}$, we denote the symmetric normalized Laplacian by $\Laplacian = \Id - \mathbf{D}^{-1/2}\mathbf{W}\mathbf{D}^{-1/2} = \Ubf\Lambdabf\Ubf^\top$, with eigenvalues $\Lambdabf=\mathrm{diag}(\lambda_1,\ldots,\lambda_C)$ where $0\le\lambda_m\le 2$.

\subsection{Definition}
\label{subsec:definition}

\paragraph{Motivation.}
The channel covariance of a brain recording is shaped in part by the sensor layout.
For example, in scalp EEG, volume conduction and shared cortical generators couple nearby electrodes~\citep{nunez2006electric}, and local cortical structure does the same for intracranial recordings.
Empirically, this makes the covariance matrix effectively low-rank.
In the eight EEG datasets we evaluate, the top-3 eigenvectors explain between $61\%$ and $100\%$ of the total channel variance, and the participation-ratio effective dimension is between $1.0$ and $7.3$ for $16$ to $64$ channels (see Table~\ref{tab:rank-def}).
An isotropic source ignores this structure, requiring the drift network to transport white noise to an effectively low-rank target.
As shown in Section~\ref{sec:analysis}, using the empirical data covariance directly degrades performance below the isotropic baseline, whereas using the eigenvectors of the position graph lowers PSD-KL.

\begin{table}[t]
\centering
\caption{Channel-covariance rank deficiency per dataset with $C$ channels.
Top-$k$ columns give the fraction of the total variance $\tr\Sbf_1$ that lies along the top-$k$ eigenvectors of the channel covariance $\Sbf_1$;
$d_{\mathrm{eff}}=(\tr\Sbf_1)^2/\tr(\Sbf_1^2)$ is the participation-ratio effective dimension. The covariance is estimated over all time samples of $512$ windows of the test split, centered per channel.}
\label{tab:rank-def}
\vspace{-0.1cm}
\adjustbox{max width=0.85\linewidth}{%
\begin{tabular}{lcrrr}
\toprule
dataset & \# channels ($C$) & top-1 & top-3 & $d_{\mathrm{eff}}(\Sbf_1)$ \\
\midrule
TUAB~\citep{obeid2016temple}         & 16 & 0.387 & 0.647 & 4.95 \\
TUEV~\citep{obeid2016temple}         & 16 & 0.330 & 0.605 & 5.87 \\
Mumtaz-MDD~\citep{mumtaz2016mdd}     & 19 & 0.997 & 0.999 & 1.01 \\
BCI-IV 2a~\citep{brunner2008bciciv2a}    & 22 & 0.494 & 0.855 & 3.02 \\
FACED~\citep{chen2023faced}        & 32 & 0.239 & 0.607 & 7.26 \\
SHU~\citep{ma2022shu}          & 32 & 0.576 & 0.816 & 2.75 \\
SEED-V~\citep{liu2022seedv}       & 62 & 0.549 & 0.796 & 2.98 \\
PhysioNet-MI~\citep{schalk2004physionet} & 64 & 0.635 & 0.775 & 2.40 \\
\bottomrule
\end{tabular}
}
\vspace{-0.3cm}
\end{table}

\paragraph{Construction.}
Given 3-D sensor coordinates $\{p_c\}_{c=1}^C$, we build an undirected $k$-nearest-neighbor graph without self-loops ($W_{ii}{=}0$), connecting sensors whenever either is among the other's $k$ nearest neighbors. Edge weights are set to $W_{ij}=\exp(-\|p_i-p_j\|^2/h^2)$, with bandwidth $h^2$ defined as the mean squared $k$-NN distance across all sensors. We then compute the normalized Laplacian $\Laplacian = \Id - \mathbf{D}^{-1/2}\mathbf{W}\mathbf{D}^{-1/2} = \Ubf\Lambdabf\Ubf^\top$ with eigenvalues $0=\lambda_1\le\lambda_2\le\cdots\le\lambda_C\le 2$ (see Figure~\ref{fig:prior-construction}).
We place an independent Gaussian on each eigenmode with variance $\phi(\lambda_m)=(1+\tau\lambda_m)^{-\alpha}$ and reassemble in sensor coordinates,
\begin{equation}
\mathrm{vec}(x_0) \;=\; (\Id_T \otimes \Ubf)\,\mathrm{vec}(\tilde x_0),\qquad \tilde x_0 \sim \N\big(0,\,\Id_T \otimes \mathrm{diag}(\phi(\Lambdabf))\big),
\label{eq:prior}
\end{equation}
where $x_0,\tilde x_0\in\R^{C\times T}$ and $\otimes$ denotes the Kronecker product.
Equivalently, $x_0 = \Ubf\,\tilde x_0$ left-multiplies each time-slice by $\Ubf$.
The $\Id_T$ in the temporal slot is a deliberate simplification, since we encode no temporal structure in the prior and leave any temporal correlations for the drift network to learn from data.
The scalar $\phi(\lambda)$ is the graph-Mat\'ern spectral density of \citet{borovitskiy2021graph}, a discrete-graph analog of the Mat\'ern Gaussian process density on Riemannian manifolds \citep{borovitskiy2020matern}, where $\tau$ sets the spatial correlation length and $\alpha$ the spectral decay rate.
We normalize $\phi$ so that $\frac{1}{C}\sum_m \phi(\lambda_m)=1$, placing the prior on the same total-variance scale as an isotropic source; the prior \emph{redistributes} variance across spatial modes.
The prior has three hyperparameters, $k$, $\tau$, and $\alpha$, whose experimental settings are described in Section~\ref{subsec:protocol}.

\paragraph{Properties.}
The prior aligns the source variance with the eigenvectors of the Laplacian of the position graph and concentrates it on the low-eigenvalue modes, lowering the prior's effective dimension $d_{\mathrm{eff}}(\Sbf_0){:=}(\tr\Sbf_0)^2/\tr(\Sbf_0^2)$ from $16$ to $10.1$ on TUAB\footnote{This is the \emph{prior}'s effective dimension, distinct from the data-covariance $d_{\mathrm{eff}}(\Sbf_1)$ in Table~\ref{tab:rank-def}.}.
When evaluated using the Tikhonov effective dimension $\tr(\Sbf_0(\Sbf_0+\lambda\Id)^{-1})$, which discounts small eigenvalues, the reduction is at most $10\%$ (see Appendix~\ref{app:pr-vs-tikhonov}); as shown in Section~\ref{sec:analysis}, the spatial eigenvectors, rather than variance concentration, drive the empirical improvement.

\begin{algorithm}[t!]
\SetAlgoLined
\KwIn{dataset $\mathcal{D}=\{x_1^{(i)}\}$, sensor coordinates $\{p_c\}_{c=1}^C$, prior hyperparameters $(\tau,\alpha,k)$, noise scale $\sigma$ with $\gamma(t)=\sigma\sqrt{t(1-t)}$, $\epsilon{=}0.001$, coupling $\pi$, drift network $v_\theta$, learning rate $\eta$, number of steps $N$.}
\KwOut{Learned drift $v_\theta^\star$.}
Build $k$-NN adjacency $\mathbf{W}$ with Gaussian weights $W_{ij}{=}\exp(-\|p_i{-}p_j\|^2/h^2)$, $h^2$ the mean squared $k$-NN distance, symmetrize, compute normalized Laplacian $\Laplacian = \Ubf\Lambdabf\Ubf^\top$; precompute $\phi(\lambda_m)=(1+\tau\lambda_m)^{-\alpha}$ normalized so $\tfrac{1}{C}\!\sum_m\phi(\lambda_m)=1$\;
\For{step $= 1,\ldots,N$}{
Sample data batch $\{x_1^{(i)}\}\sim\mathcal{D}$\;
{\color{blue}Sample prior batch $x_0 = \Ubf\,\mathrm{diag}(\sqrt{\phi(\Lambdabf)})\,z_0$, $z_0\sim\N(0,\Id_{C\times T})$ \tcp*[r]{only change}}
Couple $(x_0,x_1)\sim\pi$\;
Sample $t\sim\mathcal{U}(\epsilon,\,1{-}\epsilon)$, $z\sim\N(0,\Id)$;\quad form $X_t = (1-t)x_0 + t x_1 + \gamma(t) z$, $u_t = (x_1-x_0)+\gamma'(t)z$\;
Update $\theta \leftarrow \theta - \eta\,\nabla_\theta \|v_\theta(X_t, t) - u_t\|_2^2$\;
}
\caption{Training with the graph-Mat\'ern source prior}
\label{alg:train}
\end{algorithm}

\subsection{Composition with Flow-Matching Couplings}
\label{sec:method-objective}
The graph-Mat\'ern prior is independent of the coupling $\pi(x_0,x_1)$ and of the drift architecture.
The four methods we evaluate are listed in Section~\ref{sec:experiments}.
The training objective is the loss of Eq.~(\plaineqref{eq:fm-loss}) with the stochastic-interpolant target $u_t=(x_1-x_0)+\gamma'(t)z$ of Section~\ref{subsec:prelim}, with $\gamma(t)=\sigma\sqrt{t(1-t)}$, the vector field evaluated at the stochastic state $X_t$, and the expectation taken over $z$ as well.
Relative to an isotropic baseline, the only change is the source distribution $p_0$ (the marked line in Algorithm~\ref{alg:train}).
At inference, $X_0$ is drawn from the prior and the learned drift is integrated with $50$ Euler steps of the probability-flow ODE across all methods.
Consequently, the stochastic formulations (SI~\citep{albergo2023si}\footnote{We abbreviate the $\sigma{>}0$ \emph{stochastic interpolant} of \citet{albergo2023si} under independent coupling as SI throughout the paper.} and SF2M~\citep{tong2024sf2m}, with $\sigma{>}0$) are trained with interpolant noise and sampled deterministically without score networks (see Appendix~\ref{app:exp-details}), sharing the ODE integration scheme with the deterministic methods (OT-CFM~\citep{pooladian2023multisample,tong2024otcfm}, RF~\citep{liu2023rectified}).


\section{Experiments}
\label{sec:experiments}

\subsection{Experimental Details}\label{subsec:protocol}
\vspace{-0.2cm}

\paragraph{Methods.}
We test four flow-matching methods, each with an isotropic source ($\N(0,\Id)$) and with the graph-Mat\'ern source of Section~\ref{subsec:definition}, namely
SF2M~\citep{tong2024sf2m} (Hungarian coupling, $\sigma{>}0$)\footnote{We use the exact minibatch-OT (Hungarian) coupling on each batch of $64$ (SF2M-Exact in \citet{tong2024sf2m}), so that SF2M and OT-CFM share the coupling and differ only in $\sigma$.}, SI~\citep{albergo2023si} (independent coupling, $\sigma{>}0$), OT-CFM~\citep{pooladian2023multisample,tong2024otcfm} (Hungarian coupling, $\sigma{=}0$), and RF~\citep{liu2023rectified} (independent coupling, $\sigma{=}0$, followed by one reflow).
A method with the graph prior carries the prefix \emph{GP-} (\textit{e.g.}, GP-SI is SI with the graph-Mat\'ern source).
All methods share a 1D U-Net drift~\citep{ho2020ddpm,ronneberger2015unet} that convolves along time and treats the $C$ sensors as input channels, so the network has no spatial inductive bias and the sensor layout enters only through the source; architecture, sweep configuration, and sampler details are in Appendix~\ref{app:exp-details}.

\paragraph{Hyperparameters.}
We fix the prior hyperparameters to $k{=}4$, $\tau{=}1$, and $\alpha{=}2$ across all datasets (sensitivity analysis in Appendix~\ref{app:prior-sensitivity}).
For methods with a stochastic bridge ($\sigma{>}0$), we use noise scale $\sigma{=}0.02$, adjusted to $\sigma{=}0.008$ on TUEV to account for its $2.5\times$ smaller signal amplitude, which maintains $\sigma/\sigma_{\mathrm{real}} \approx 0.08$ relative to the per-channel training standard deviation $\sigma_{\mathrm{real}}$ on every dataset except FACED, where it is $0.19$ (see Appendix~\ref{app:sigma-sweep}).

\begin{table}[t!]
\centering
\caption{PSD-KL on eight EEG datasets (test split, mean$\pm$std over five training seeds; lower is better) for the four flow-matching methods with an isotropic source and with the graph-Mat\'ern source (GP-).
Bold denotes the better source within each method.
Because PSD-KL is not comparable across datasets, the last row gives the geometric mean over datasets of the ratio GP/isotropic (below one favors the graph source).
}
\label{tab:sweep-all-main}
\setlength{\tabcolsep}{2pt}
\adjustbox{max width=\linewidth}{%
\begin{tabular}{l cc cc cc cc}
\toprule
& \multicolumn{2}{c}{$\sigma{>}0$ Hungarian} & \multicolumn{2}{c}{$\sigma{>}0$ independent} & \multicolumn{2}{c}{$\sigma{=}0$ Hungarian} & \multicolumn{2}{c}{$\sigma{=}0$ indep+reflow} \\
\cmidrule(lr){2-3}\cmidrule(lr){4-5}\cmidrule(lr){6-7}\cmidrule(lr){8-9}
dataset (ch) & SF2M & GP-SF2M & SI & GP-SI & OT-CFM & GP-OT-CFM & RF & GP-RF \\
\midrule
TUAB (16)        & $1.68\!\pm\!0.23$ & $\mathbf{1.04\!\pm\!0.28}$ & $1.43\!\pm\!0.37$ & $\mathbf{1.16\!\pm\!0.20}$ & $1.67\!\pm\!0.33$ & $\mathbf{1.40\!\pm\!0.38}$ & $\mathbf{2.27\!\pm\!1.70}$ & $2.53\!\pm\!1.13$ \\
TUEV (16)        & $\mathbf{2.03\!\pm\!0.33}$ & $3.85\!\pm\!2.47$ & $2.55\!\pm\!0.78$ & $\mathbf{2.15\!\pm\!1.10}$ & $5.30\!\pm\!2.30$ & $\mathbf{3.80\!\pm\!0.59}$ & $3.19\!\pm\!1.21$ & $\mathbf{3.09\!\pm\!0.29}$ \\
Mumtaz (19)      & $4.42\!\pm\!3.33$ & $\mathbf{3.89\!\pm\!1.94}$ & $3.91\!\pm\!2.42$ & $3.91\!\pm\!1.11$ & $\mathbf{4.97\!\pm\!2.06}$ & $5.51\!\pm\!1.66$ & $6.07\!\pm\!3.02$ & $\mathbf{3.79\!\pm\!1.33}$ \\
BCI-IV 2a (22)   & $3.78\!\pm\!1.00$ & $\mathbf{3.56\!\pm\!1.47}$ & $\mathbf{3.30\!\pm\!0.73}$ & $3.50\!\pm\!0.59$ & $\mathbf{3.06\!\pm\!0.73}$ & $3.61\!\pm\!0.19$ & $4.27\!\pm\!0.71$ & $\mathbf{4.24\!\pm\!1.35}$ \\
FACED (32)       & $32.86\!\pm\!4.01$ & $\mathbf{31.77\!\pm\!7.76}$ & $33.34\!\pm\!3.54$ & $\mathbf{29.03\!\pm\!4.56}$ & $33.60\!\pm\!7.21$ & $\mathbf{31.48\!\pm\!12.05}$ & $32.14\!\pm\!3.91$ & $\mathbf{26.44\!\pm\!4.96}$ \\
SHU (32)         & $113.08\!\pm\!3.90$ & $\mathbf{81.36\!\pm\!8.01}$ & $114.36\!\pm\!1.73$ & $\mathbf{82.76\!\pm\!4.57}$ & $111.27\!\pm\!5.19$ & $\mathbf{85.36\!\pm\!0.71}$ & $115.72\!\pm\!2.26$ & $\mathbf{82.88\!\pm\!10.33}$ \\
SEED-V (62)      & $32.84\!\pm\!0.75$ & $\mathbf{28.05\!\pm\!0.19}$ & $32.93\!\pm\!0.71$ & $\mathbf{28.02\!\pm\!0.13}$ & $32.90\!\pm\!0.73$ & $\mathbf{28.06\!\pm\!0.31}$ & $33.33\!\pm\!0.38$ & $\mathbf{28.26\!\pm\!0.21}$ \\
PhysioNet-MI (64) & $34.78\!\pm\!0.88$ & $\mathbf{20.52\!\pm\!3.67}$ & $33.42\!\pm\!0.88$ & $\mathbf{19.60\!\pm\!2.37}$ & $32.98\!\pm\!1.37$ & $\mathbf{19.87\!\pm\!3.90}$ & $33.34\!\pm\!1.83$ & $\mathbf{22.44\!\pm\!1.54}$ \\
\midrule
\textbf{geometric mean of GP/iso} & \multicolumn{2}{c}{$0.88$} & \multicolumn{2}{c}{$0.83$} & \multicolumn{2}{c}{$0.86$} & \multicolumn{2}{c}{$0.83$} \\
\bottomrule
\end{tabular}}
\end{table}

\paragraph{Metrics.}
Our primary metric, PSD-KL, fits a Gaussian to the log band power across windows for each channel and clinical band ($\delta, \theta, \alpha, \beta, \gamma$) and reports the symmetric KL divergence between the real and generated fits, averaged over channels and bands (lower is better; see Appendix~\ref{app:exp-details}).
Because the divergence scales with each dataset's empirical variance of log band power and diverges under variance mismatch, absolute values are not comparable across datasets; comparisons are made within each dataset.
Band power is the standard summary of an EEG recording~\citep{nunez2006electric}, whereas general-purpose scores like the Fr\'echet Inception Distance~\citep{heusel2017fid} can disagree with spectral fidelity on EEG, where the model with the most realistic spatial and spectral properties was assigned the worst FID~\citep{hartmann2018eeggan}.
To complement this per-channel metric, we quantify cross-channel coupling using the weighted phase-lag index (wPLI)~\citep{vinck2011wpli}, evaluated as the Pearson correlation between real and generated wPLI matrices (higher is better).

\paragraph{Data and statistics.}
We evaluate on eight EEG datasets (TUAB and TUEV~\citep{obeid2016temple}, Mumtaz-MDD~\citep{mumtaz2016mdd}, BCI-IV 2a~\citep{brunner2008bciciv2a}, FACED~\citep{chen2023faced}, SHU~\citep{ma2022shu}, SEED-V~\citep{liu2022seedv}, and PhysioNet-MI~\citep{schalk2004physionet}) spanning $16$ to $64$ channels across resting-state, event-related, motor-imagery, and emotion paradigms.
Preprocessing and data splits follow \citet{wang2024cbramod}. All metrics are evaluated on the official test split, which holds out subjects on six of the eight datasets (SEED-V partitions trials within session, while SHU shares sessions across splits; see Appendix~\ref{app:exp-details}).
All experiments use the same five training seeds, shared across both sources; reported values indicate the mean $\pm$ standard deviation over seeds.
The MEG, intracranial EEG, and traffic experiments of Section~\ref{sec:ieeg} use the same methods, sources, seeds, and metrics; their graphs, training lengths, and splits are in Appendix~\ref{app:ieeg-per-patient}.

\subsection{The Graph Prior Improves the Power Spectrum on EEG}
\label{subsec:cross-dataset}

\begin{figure}[t]
\centering
\includegraphics[width=0.85\linewidth]{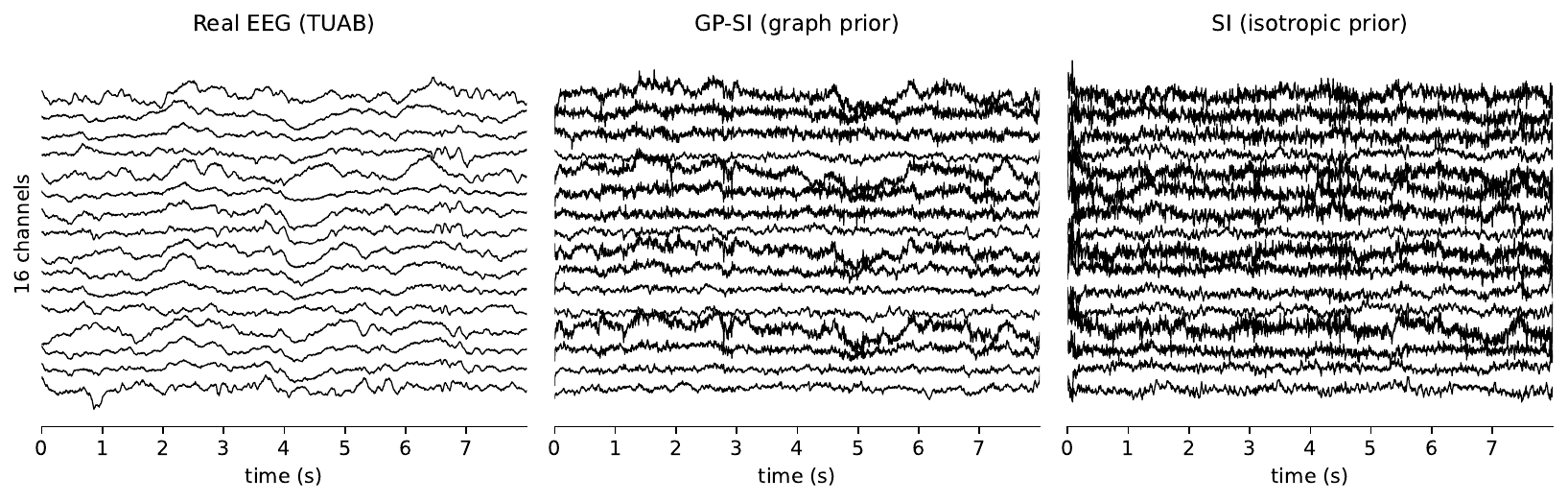}
\vspace{-0.2cm}
\caption{\textbf{Qualitative effect of the graph prior (GP) on TUAB~\citep{obeid2016temple}.}
16-channel waveforms from one real test window and one sample each from GP-SI (SI trained with the graph prior) and SI~\citep{albergo2023si}.
Real EEG shows slow oscillations that are coherent across channels; GP-SI samples retain both properties, whereas SI samples resemble channel-independent high-frequency noise.}
\label{fig:qual_tuab}
\vspace{-0.3cm}
\end{figure}

\paragraph{Power spectrum.}
As shown in Table~\ref{tab:sweep-all-main}, the graph prior lowers PSD-KL across all four flow-matching formulations, achieving geometric mean ratios of $0.83$--$0.88$ relative to the isotropic baseline.
For every method, it lowers PSD-KL on most datasets, and the exceptions are discussed at the end of this section.
The largest reductions shared by all four methods occur on PhysioNet-MI (64 channels), where PSD-KL drops from $33$--$35$ to $20$--$22$, and on SHU (32 channels), where it drops from $111$--$116$ to $81$--$85$.
A one-sided Wilcoxon signed-rank test on $\log(\mathrm{iso}/\mathrm{graph})$ across the eight datasets yields uncorrected $p$-values of $0.016$ for SI, $0.020$ for RF, $0.055$ for OT-CFM, and $0.098$ for SF2M (adjusted to $0.063$ for SI and RF after Holm correction).
Figure~\ref{fig:qual_tuab} shows one example from TUAB.
GP-SI samples retain the slow oscillations and inter-channel coherence of real EEG, whereas SI samples resemble channel-independent high-frequency noise.

\paragraph{Cross-channel coupling.}
The graph prior raises the wPLI correlation on average for all four methods by $0.01$--$0.03$ as shown in Table~\ref{tab:wpli-main}, with consistent gains across all four methods on the densest montages, SEED-V and PhysioNet-MI ($+0.05$--$0.12$).
On the remaining datasets, differences generally remain within standard deviations, except on TUAB, where GP-SI improves and GP-OT-CFM decreases by $0.06$, and on FACED, where GP-RF decreases by $0.05$.
Across the eight datasets, these differences do not reach statistical significance after Holm correction.
Because the graph prior directly shapes zero-lag spatial covariance, whereas wPLI explicitly discards zero-lag interactions to eliminate volume conduction, the small average effect on this metric ($0.01$ to $0.03$) is expected.

\begin{table}[t]
\centering
\caption{wPLI correlation between real and generated data on the same eight datasets (band-averaged correlation, mean$\pm$std over five training seeds; higher is better) for the four methods with an isotropic and with the graph-Mat\'ern source.
Bold denotes the better source within each method.
}
\label{tab:wpli-main}
\setlength{\tabcolsep}{3pt}
\adjustbox{max width=\linewidth}{%
\begin{tabular}{l cc cc cc cc}
\toprule
& \multicolumn{2}{c}{$\sigma{>}0$ Hungarian} & \multicolumn{2}{c}{$\sigma{>}0$ independent} & \multicolumn{2}{c}{$\sigma{=}0$ Hungarian} & \multicolumn{2}{c}{$\sigma{=}0$ indep+reflow} \\
\cmidrule(lr){2-3}\cmidrule(lr){4-5}\cmidrule(lr){6-7}\cmidrule(lr){8-9}
dataset (ch) & SF2M & GP-SF2M & SI & GP-SI & OT-CFM & GP-OT-CFM & RF & GP-RF \\
\midrule
TUAB (16)        & $\mathbf{0.11\!\pm\!0.10}$ & $0.06\!\pm\!0.12$ & $0.02\!\pm\!0.05$ & $\mathbf{0.08\!\pm\!0.05}$ & $\mathbf{0.13\!\pm\!0.05}$ & $0.07\!\pm\!0.05$ & $0.03\!\pm\!0.04$ & $\mathbf{0.05\!\pm\!0.05}$ \\
TUEV (16)        & $-0.00\!\pm\!0.03$ & $\mathbf{0.01\!\pm\!0.04}$ & $\mathbf{0.03\!\pm\!0.02}$ & $0.00\!\pm\!0.05$ & $-0.03\!\pm\!0.04$ & $\mathbf{-0.01\!\pm\!0.07}$ & $0.01\!\pm\!0.02$ & $0.01\!\pm\!0.05$ \\
Mumtaz (19)      & $0.31\!\pm\!0.16$ & $\mathbf{0.41\!\pm\!0.06}$ & $\mathbf{0.40\!\pm\!0.03}$ & $0.39\!\pm\!0.02$ & $0.39\!\pm\!0.06$ & $\mathbf{0.40\!\pm\!0.03}$ & $0.33\!\pm\!0.13$ & $\mathbf{0.38\!\pm\!0.07}$ \\
BCI-IV 2a (22)   & $0.19\!\pm\!0.05$ & $\mathbf{0.22\!\pm\!0.07}$ & $0.19\!\pm\!0.08$ & $\mathbf{0.22\!\pm\!0.06}$ & $0.17\!\pm\!0.08$ & $\mathbf{0.23\!\pm\!0.06}$ & $\mathbf{0.17\!\pm\!0.04}$ & $0.13\!\pm\!0.03$ \\
FACED (32)       & $0.04\!\pm\!0.05$ & $\mathbf{0.06\!\pm\!0.07}$ & $0.08\!\pm\!0.03$ & $\mathbf{0.09\!\pm\!0.06}$ & $0.06\!\pm\!0.03$ & $\mathbf{0.10\!\pm\!0.05}$ & $\mathbf{0.12\!\pm\!0.02}$ & $0.07\!\pm\!0.02$ \\
SHU (32)         & $\mathbf{0.03\!\pm\!0.04}$ & $0.01\!\pm\!0.04$ & $\mathbf{0.02\!\pm\!0.04}$ & $0.01\!\pm\!0.02$ & $\mathbf{0.03\!\pm\!0.05}$ & $0.01\!\pm\!0.02$ & $0.01\!\pm\!0.01$ & $\mathbf{0.02\!\pm\!0.03}$ \\
SEED-V (62)      & $-0.01\!\pm\!0.01$ & $\mathbf{0.06\!\pm\!0.02}$ & $-0.01\!\pm\!0.02$ & $\mathbf{0.06\!\pm\!0.01}$ & $-0.01\!\pm\!0.02$ & $\mathbf{0.06\!\pm\!0.03}$ & $-0.02\!\pm\!0.02$ & $\mathbf{0.03\!\pm\!0.04}$ \\
PhysioNet-MI (64) & $0.33\!\pm\!0.05$ & $\mathbf{0.40\!\pm\!0.04}$ & $0.32\!\pm\!0.05$ & $\mathbf{0.44\!\pm\!0.02}$ & $0.30\!\pm\!0.04$ & $\mathbf{0.42\!\pm\!0.06}$ & $0.29\!\pm\!0.03$ & $\mathbf{0.38\!\pm\!0.06}$ \\
\midrule
\textbf{mean across datasets} & $0.12$ & $\mathbf{0.15}$ & $0.13$ & $\mathbf{0.16}$ & $0.13$ & $\mathbf{0.16}$ & $0.12$ & $\mathbf{0.13}$ \\
\bottomrule
\end{tabular}}
\vspace{-0.3cm}
\end{table}

\paragraph{Datasets without a gain.}
Each case in Table~\ref{tab:sweep-all-main} where the graph prior does not improve on isotropic noise has a likely explanation.
On BCI-IV 2a, the 22 electrodes cover only the centro-parietal region, so the graph has little spatial structure to encode, whereas a similar motor-imagery task recorded with 64 electrodes over the whole scalp (PhysioNet-MI) gives our largest overall gain.
On FACED, the prior lowers PSD-KL for all four methods at the default noise scale, and only at a five times larger one ($\sigma{=}0.10$) is it worse in $\beta$ and $\gamma$ while matching or improving the lower bands.
However, because the prior lowers the $\gamma$-band error on SHU from $340$ to $261$ for SI, this behavior likely reflects dataset-specific high-frequency noise rather than an inherent limit of spatial priors.
On TUEV, only SF2M is worse with the prior, and its increase is spread over all five frequency bands.
On Mumtaz, SI is unchanged and OT-CFM is $11\%$ worse, and on TUAB, RF is $11\%$ worse, with these differences remaining within seed standard deviations.
Appendix~\ref{app:sweep-all} gives the per-band results behind the FACED, SHU, and TUEV statements.

\subsection{Generalization to MEG, Intracranial EEG, and Traffic Sensors}
\label{sec:ieeg}
\label{subsec:ieeg}
To test whether the graph prior applies beyond multi-channel scalp EEG, we evaluate the identical construction on other spatial sensor systems using coordinates alone (see Appendix~\ref{app:ieeg-per-patient}).
As shown in Table~\ref{tab:cross-modality}, the graph prior lowers PSD-KL across all three domains, whole-head MEG (Wakeman--Henson~\citep{wakemanhenson2015}; $16$ subjects, $102$ magnetometers on a shared helmet), intracranial EEG (CCEP ECoG~\citep{vanblooijs2023ccep}; five patients with individual $72$--$112$-channel ECoG grids), and the PEMS-BAY traffic network~\citep{li2018dcrnn} ($325$ freeway loop sensors, graph from GPS coordinates).

\begin{table}[t!]
\centering
\caption{PSD-KL on MEG, intracranial EEG, and a traffic-sensor network (lower is better; mean over five seeds and, for MEG and iEEG, over subjects or patients) for the four methods with an isotropic and with the graph-Mat\'ern source.
Bold denotes the better source within each method.
The graph prior lowers PSD-KL across all four methods and all three modalities (one-sided sign test, $p < 0.0001$ over MEG subjects and $p = 0.031$ over iEEG patients and over traffic seeds).
Absolute values are not comparable across rows; per-patient iEEG values are in Table~\ref{tab:ieeg-per-patient}.
}
\vspace{-0.1cm}
\label{tab:cross-modality}
\setlength{\tabcolsep}{4pt}
\adjustbox{max width=\linewidth}{%
\begin{tabular}{l cc cc cc cc}
\toprule
& \multicolumn{2}{c}{$\sigma{>}0$ Hungarian} & \multicolumn{2}{c}{$\sigma{>}0$ independent} & \multicolumn{2}{c}{$\sigma{=}0$ Hungarian} & \multicolumn{2}{c}{$\sigma{=}0$ indep+reflow} \\
\cmidrule(lr){2-3}\cmidrule(lr){4-5}\cmidrule(lr){6-7}\cmidrule(lr){8-9}
modality & SF2M & GP-SF2M & SI & GP-SI & OT-CFM & GP-OT-CFM & RF & GP-RF \\
\midrule
MEG, Wakeman--Henson ($16$ subjects)   & 2890 & \textbf{2715} & 2891 & \textbf{2715} & 2893 & \textbf{2712} & 2902 & \textbf{2726} \\
iEEG, CCEP ECoG ($5$ patients)    & 31.3 & \textbf{19.2} & 31.2 & \textbf{18.7} & 30.5 & \textbf{18.9} & 37.6 & \textbf{24.9} \\
PEMS-BAY traffic ($325$ sensors)       & 62.1 & \textbf{55.6} & 62.1 & \textbf{55.5} & 62.1 & \textbf{55.3} & 63.9 & \textbf{56.8} \\
\bottomrule
\end{tabular}}
\vspace{-0.2cm}
\end{table}

The relative reduction is largest on iEEG ($34$--$40\%$), where each patient has a distinct cortical grid, intermediate on the traffic network ($11\%$), and smallest on MEG ($6\%$), where all subjects share a fixed sensor geometry.
Unlike neural recordings, the freeway network involves no biological tissue or volume conduction.
The gain on this network therefore cannot come from properties specific to electrophysiology.

\begin{table}[t!]
\centering
\caption{Downstream augmentation on Mumtaz-MDD with an EEGNet-8,2 classifier (subject-disjoint split with $31$ training, $6$ validation, and $8$ test subjects; every column is the mean$\pm$std over five seeds).
Rows add one generated window per real training window from the isotropic or the graph-prior source, each generated window first matched in amplitude to the real training windows (see Appendix~\ref{app:exp-details}).
Bold denotes the better source.
}
\label{tab:downstream-mumtaz}
\vspace{-0.1cm}
\adjustbox{max width=0.85\linewidth}{%
\begin{tabular}{lccc}
\toprule
method & balanced accuracy & recall, minority class & recall, majority class \\
\midrule
real only                & $0.608 \pm 0.146$ & $0.284 \pm 0.386$ & $0.931 \pm 0.096$ \\
class-weighted real only & $0.545 \pm 0.097$ & $0.091 \pm 0.194$ & $0.999 \pm 0.002$ \\
$+$ iso-source           & $0.819 \pm 0.037$ & $0.786 \pm 0.009$ & $0.852 \pm 0.075$ \\
$+$ graph-prior          & $\mathbf{0.833 \pm 0.054}$ & $\mathbf{0.794 \pm 0.006}$ & $\mathbf{0.872 \pm 0.110}$ \\
\bottomrule
\end{tabular}}
\vspace{-0.3cm}
\end{table}

\subsection{Data Augmentation for EEG Classification}
\label{sec:downstream}

To evaluate whether generative improvements translate to downstream utility, we benchmark data augmentation on Mumtaz-MDD, a binary depression-detection task with a subject-disjoint split ($31$ training, $6$ validation, and $8$ test subjects, $841$ test windows).
We train class-conditional generators on the training split only and train an EEGNet-8,2 classifier~\citep{lawhern2018eegnet} on real windows with and without synthetic data added.
Before augmenting the training set, generated windows are matched in root-mean-square amplitude to the real training data (see Appendix~\ref{app:exp-details}).
As shown in Table~\ref{tab:downstream-mumtaz}, synthetic windows increase balanced accuracy from $60.8\%$ to $83.3\%$ with the graph prior and to $81.9\%$ with the isotropic baseline.
This gain is driven primarily by the minority class, where recall rises from $0.284$ to $0.794$.
The improvement is not achievable by loss reweighting alone, as class-weighted training on real data reaches only $54.5\%$ balanced accuracy.


\section{Which Part of the Graph Prior Carries the Gain?}
\label{sec:analysis}

\begin{table}[t]
\centering
\caption{Ablating the graph-Mat\'ern prior on TUAB with SI.
PSD-KL on the validation split at the configuration of TUAB's row in Table~\ref{tab:sweep-all-main}, mean$\pm$std over five training seeds, over $25$ random permutations for the shuffled-position row, and over ten seeds for the correlation-graph row; the reference rows therefore differ from the test-split values of that row.
The second column names the part of the construction each row replaces, $d_{\mathrm{eff}}$ is the prior's effective dimension, and bold denotes the best prior (details in Appendix~\ref{app:mechanism}).}
\vspace{-0.1cm}
\label{tab:prior-ablation}
\adjustbox{max width=0.85\linewidth}{%
\begin{tabular}{llcc}
\toprule
prior & changes & $d_{\mathrm{eff}}$ & PSD-KL \\
\midrule
isotropic                         & ---                       & $16.0$ & $1.85 \pm 0.18$ \\
graph-Mat\'ern (graph of sensor positions) & ---                       & $10.1$ & $\mathbf{1.08 \pm 0.41}$ \\
\midrule
shuffled sensor positions         & eigenvectors              & $10.1$ & $3.70 \pm 2.79$ \\
uniformly random eigenvectors     & eigenvectors              & $10.1$ & $4.42 \pm 2.97$ \\
empirical eigenvectors, graph spectrum & eigenvectors           & $10.1$ & $2.76 \pm 2.93$ \\
empirical covariance              & eigenvectors and spectrum & $5.2$  & $4.61 \pm 2.40$ \\
Ledoit--Wolf-shrunk empirical covariance & eigenvectors and spectrum & $6.6$ & $3.56 \pm 2.43$ \\
\midrule
heat-kernel spectrum              & spectrum                  & $12.0$ & $1.38 \pm 0.86$ \\
hard low-pass spectrum            & spectrum                  & $8.0$  & $71.40 \pm 12.50$ \\
\midrule
fully connected Gaussian graph    & graph                     & $11.2$ & $2.53 \pm 1.81$ \\
correlation graph ($k$-NN on $|\mathrm{corr}|$) & graph      & $11.0$ & $1.39 \pm 1.70$ \\
\bottomrule
\end{tabular}}
\vspace{-0.4cm}
\end{table}

We next analyze which properties of the graph prior account for its empirical gains.
The graph-Mat\'ern source differs from isotropic noise in three structural aspects.
Its variance is concentrated on fewer modes (effective dimension $10.1$ vs.\ $16.0$ on TUAB), its spectrum decays smoothly along the Laplacian eigenvalues, and its eigenvectors are defined by the graph of sensor positions.
Any of these factors could explain the observed improvement.
We therefore train SI on TUAB, the simplest of the four methods, with priors that replace the eigenvectors, the spectrum, or the graph while keeping the rest fixed, and compare them in Table~\ref{tab:prior-ablation} against the isotropic source (PSD-KL $1.85 \pm 0.18$) and the graph prior ($1.08 \pm 0.41$).

\paragraph{The gain requires the physical sensor eigenbasis.}
To test whether the improvement stems from sensor geometry or merely from low effective dimension, we keep the Mat\'ern spectrum ($d_{\mathrm{eff}}{=}10.1$) but substitute the eigenvectors (see Appendix~\ref{app:random-eigenbasis}).
Replacing them with permuted sensor positions ($3.70 \pm 2.79$ over $25$ permutations), a uniformly random orthonormal basis ($4.42 \pm 2.97$), or empirical covariance eigenvectors ($2.76 \pm 2.93$) is worse than the isotropic baseline ($1.85 \pm 0.18$) and raises the standard deviation over seeds about sevenfold relative to the graph prior ($0.41$).
An isotropic prior is at least spatially neutral, whereas any other eigenbasis imposes correlations that the drift must undo.
Spectral concentration is therefore beneficial only when guided by the physical sensor layout.

\paragraph{Concentrating variance in the wrong directions is worse than no structure.}
Using the empirical channel covariance itself as the prior combines eigenvectors estimated from the data with a more concentrated spectrum ($d_{\mathrm{eff}}{=}5.2$).
At $4.61 \pm 2.40$, this prior underperforms all priors that replace only the eigenvectors, so concentrating variance helps only along the eigenvectors of the position graph.
Shrinking the empirical covariance toward a multiple of the identity with the Ledoit--Wolf estimator~\citep{ledoit2004well}, which keeps its eigenvectors and only flattens its spectrum, is also worse than the isotropic source ($3.56 \pm 2.43$ against $1.85 \pm 0.18$; see Appendix~\ref{app:lw-prior}).
This failure is not caused by near-zero eigenvalues in the empirical covariance, since the prior that keeps empirical eigenvectors but uses the graph's Mat\'ern spectrum, which has no small eigenvalues, also underperforms the isotropic source ($2.76 \pm 2.93$).
Nor can the failure be explained by Gaussian source--target covariance mismatch, as on a synthetic Gaussian target matching the EEG channel covariance, all six tested Gaussian sources reach a similar PSD-KL ($0.0020$--$0.0032$; see Appendix~\ref{app:learned-prior}).
The gap on real EEG therefore comes from its non-Gaussianity and its temporal correlations, which this Gaussian target lacks.

\paragraph{The spectrum only needs to be smooth and to cover every mode.}
Replacing the Mat\'ern density with a heat kernel $\exp(-\tau\lambda)$ on the same eigenvectors keeps most of the gain ($1.38 \pm 0.86$; see Appendix~\ref{app:spectrum-shape}).
A hard low-pass that zeroes the upper half of the modes fails ($71.40 \pm 12.50$), since the flow is a smooth invertible map and cannot spread a source with no variance on half of the modes over a target that has variance on all of them; the drift has to create all of the missing variance.

\paragraph{The graph must be sparse and local.}
Connecting all sensor pairs with a fully connected Gaussian graph rather than restricting edges to $k$ nearest neighbors yields $2.53 \pm 1.81$, performing worse than isotropic noise ($1.85 \pm 0.18$).
Sensor geometry alone is therefore insufficient; the inductive bias requires the eigenvectors of a sparse local graph.
A $k$-NN graph built from channel correlations instead of positions yields $1.39 \pm 1.70$ across ten seeds (median $0.90$); while it is competitive on most seeds, its standard deviation over seeds is four times that of the graph prior, likely because data-driven correlations inherit recording artifacts absent in physical coordinates (see Appendix~\ref{app:prior-structure}).
In summary, these experiments indicate that the gain is carried by the eigenvectors of a sparse local graph over the sensors.
The spectrum needs only to be smooth and full-rank, and sensor coordinates provide such a graph without access to the data.
Two baselines that instead whiten the data with the empirical channel covariance (ZCA whitening) or put the graph into the drift network (a graph-convolution mixer) perform within the standard deviation of the isotropic baseline at this configuration (see Appendix~\ref{app:precond-baselines}).


\section{Related Work}
\label{sec:related}

\paragraph{Structured sources and geometric priors in generative models.}
Non-isotropic noise schedules in diffusion alter the covariance profile along the sampling trajectory while retaining an isotropic base distribution~\citep{dockhorn2022cld,hoogeboom2023blurring,shaul2023kinetic}.
Similarly, geometric generative models build physical constraints into the drift or score network~\citep{de2022riemannian,dutordoir2023neural,hoogeboom2022equivariant}.
Few methods change the base distribution itself.
PriorGrad~\citep{lee2022priorgrad} adapts diagonal covariance matrices to conditioning inputs in speech synthesis, non-isotropic~\citep{voleti2022cold} and function-space diffusions~\citep{kerrigan2023infinite,pidstrigach2024infinite} define Gaussian-process priors over continuous domains, and graph-aware diffusion~\citep{rozada2026graph} obtains a graph-Laplacian covariance from heat diffusion in the forward process.
Within flow matching, TSFlow~\citep{kollovieh2025tsflow} structures the source along the temporal axis with a Gaussian process for time-series forecasting.
In contrast, our approach leaves the flow-matching objective, couplings, and drift parameterization unchanged and uses the graph-Mat\'ern spectral density~\citep{borovitskiy2021graph}, the discrete analog of manifold kernels~\citep{borovitskiy2020matern}, as a spatial source covariance that can be combined with a temporal prior as a Kronecker product (see Appendix~\ref{app:temporal-source}).

\paragraph{Generative modeling of multi-channel electrophysiology.}
Generative modeling of electroencephalography (EEG) spans generative adversarial networks~\citep{hartmann2018eeggan}, as well as diffusion models for decoding imagined speech~\citep{kim2023diffe} and for synthesizing event-related potentials~\citep{klein2024diffusionerp}.
While discriminative neural architectures build spatial structure into the network through learned spatial filters~\citep{lawhern2018eegnet} or per-electrode embeddings~\citep{jiang2024labram}, generative models still start from an isotropic Gaussian.
The generator, therefore, has to learn volume conduction and cross-channel covariance from the data alone.
Building the source from the graph of sensor positions builds this structure into the prior, independently of the architecture and for any sensor layout.

\section{Discussion}
\label{sec:discussion}

We have introduced a graph-Mat\'ern source prior for flow-matching models of multi-channel brain signals, constructed from the sensor coordinates alone, without learned parameters.
Across four flow-matching formulations and eight EEG datasets, replacing the standard isotropic Gaussian source with the graph prior reduces spectral divergence (PSD-KL) by $12\%$ to $17\%$ in geometric mean across datasets and up to $40\%$ on PhysioNet-MI, while raising the correlation between real and generated phase-lag coupling (wPLI) by $0.01$ to $0.03$ on average.
The same prior also lowers PSD-KL on whole-head MEG, intracranial EEG with patient-specific grids, and a freeway traffic-sensor network.
Ablation experiments on TUAB indicate that the gain comes from the eigenvectors of a sparse local graph over sensor coordinates, rather than from variance concentration or the parametric decay of the Mat\'ern spectrum.
Priors estimated from the empirical covariance underperform isotropic noise even after shrinkage regularization.

The prior shapes only the spatial covariance and leaves all temporal structure to the velocity network.
A separable spatiotemporal prior with a temporal Gaussian process~\citep{kollovieh2025tsflow} is a natural extension.
Our evaluation measures spectral match and phase coupling rather than overall sample realism, and PSD-KL compares sources within a dataset, not across datasets.
Position graphs suffice for scalp and grid recordings, and manifold Mat\'ern kernels~\citep{borovitskiy2020matern} could extend the prior to folded cortical surface meshes in source-space analyses.
Finally, amplitude-matched synthetic windows raise downstream balanced accuracy on Mumtaz-MDD from $60.8\%$ to $82$--$83\%$, and whether geometry-based sources help on other clinical tasks and in continuous-time diffusion models~\citep{rozada2026graph} remains to be tested.

\bibliography{references}
\bibliographystyle{iclr2027_conference}

\appendix

\section{Experimental Details}
\label{app:exp-details}

\paragraph{Graph construction.}
Sensor coordinates are scaled so that the largest radius is one before the graph is built.
The edge-weight bandwidth $h^2$ is one global value, the mean over all sensors of the squared distance to their $k$ nearest-neighbors, so neither the $k$-NN graph nor the weights depend on the units of the coordinates (EEG head frame, MEG helmet, ECoG grid in millimeters, or Earth-centered traffic coordinates).
A single bandwidth treats a dense and a sparse region of an irregular layout, such as an ECoG grid, with the same scale; a per-sensor bandwidth would adapt to local spacing, which we leave for future exploration.

\paragraph{Sampling.}
In the EEG experiments, sampling integrates the learned velocity with $50$ Euler steps of the probability-flow ODE from $t{=}0$ to $1$ for every method, including the $\sigma{>}0$ ones, and no score network is trained; the $\sigma{>}0$ models are therefore trained with the interpolant noise and sampled without it, identically for both sources, so the sampler is not the bridge SDE of \citet{tong2024sf2m}.
Sampling instead with $\sigma\sqrt{\Delta t}$ Gaussian noise added after each Euler step, on the runs of Appendix~\ref{app:temporal-source}, changes PSD-KL by less than $6\%$ for every dataset-method pair of SI and SF2M on the seven datasets of Table~\ref{tab:temporal-source} except Mumtaz with GP-SF2M ($13\%$) and never alters the relative ranking of the two priors, indicating that conclusions are robust to the choice of sampler.

\paragraph{Datasets.}
Table~\ref{tab:datasets} summarizes the eight EEG datasets of the sweep.
Preprocessing, montages, and splits follow \citet{wang2024cbramod}; all recordings are resampled to $200$\,Hz, and each run trains on at most $20{,}000$ windows of the training split, using all available windows when fewer exist.
Subject and window counts are those of the copies and splits we use.
The test split holds out subjects on TUAB, TUEV, Mumtaz-MDD, BCI-IV 2a, FACED, and PhysioNet-MI; on SEED-V the copy we use covers subjects 1--3 (seven sessions) and are split by trial within each session (trials 1--5, 6--10, and 11--15 for training, validation, and test), so every subject appears in every split; on SHU the released test split shares $30$ subject-sessions with the training split, including $1{,}785$ identical windows, so neither dataset is subject-held-out and their test-split numbers should be read as within-subject fits.
The Mumtaz-MDD recordings we use cover $30$ patients and $15$ controls, split $31$/$6$/$8$ by subject; BCI-IV 2a uses the training session of each of the nine subjects.

\begin{table}[h]
\centering
\caption{The eight EEG datasets of the sweep. Windows are given in seconds and, in parentheses, in samples at $200$\,Hz; the native rate is the rate of the released recordings; window counts are for the splits we use before the $20{,}000$-window training cap.}
\label{tab:datasets}
\vspace{0.1cm}
\adjustbox{max width=\linewidth}{%
\begin{tabular}{llrlrrr}
\toprule
dataset & recording & subjects & channels (montage) & window & native rate & windows \\
\midrule
TUAB~\citep{obeid2016temple} & clinical, normal vs.\ abnormal & $2{,}993$ (recordings) & $16$ (bipolar longitudinal) & $10$\,s ($2000$) & $256$\,Hz & $409{,}455$ \\
TUEV~\citep{obeid2016temple} & clinical events, six classes & --- & $16$ (bipolar longitudinal) & $5$\,s ($1000$) & $256$\,Hz & $113{,}353$ \\
Mumtaz-MDD~\citep{mumtaz2016mdd} & depression vs.\ control, rest & $45$ & $19$ (10--20) & $5$\,s ($1000$) & $256$\,Hz & $5{,}087$ \\
BCI-IV 2a~\citep{brunner2008bciciv2a} & motor imagery, four classes & $9$ & $22$ (centro-parietal) & $4$\,s ($800$) & $250$\,Hz & $2{,}592$ \\
FACED~\citep{chen2023faced} & emotion, video clips & $123$ & $32$ (10--20) & $10$\,s ($2000$) & $250$\,Hz & $10{,}332$ \\
SHU~\citep{ma2022shu} & motor imagery, two classes & $52$ & $32$ (10--20) & $4$\,s ($800$) & $250$\,Hz & $29{,}994$ \\
SEED-V~\citep{liu2022seedv} & emotion, five classes & $3$ & $62$ (10--20) & $1$\,s ($200$) & $1000$\,Hz & $17{,}461$ \\
PhysioNet-MI~\citep{schalk2004physionet} & motor imagery, four classes & $109$ & $64$ (10--10) & $4$\,s ($800$) & $160$\,Hz & $9{,}837$ \\
\bottomrule
\end{tabular}}
\end{table}

\paragraph{Hyperparameters.}
All runs use AdamW with learning rate $0.0002$, weight decay $0.00001$, gradient-norm clipping at $1.0$, and batch size $64$; the sweep and the ablation study keeps the learning rate constant, and only the sensitivity sweep of Appendix~\ref{app:prior-sensitivity} and the MEG, iEEG, and PEMS-BAY runs of Appendix~\ref{app:ieeg-per-patient} use a cosine schedule.
The eight-dataset sweep uses a U-Net with base channel width $48$ ($8.4$M parameters) trained for $6$k steps; TUAB's row of Table~\ref{tab:sweep-all-main} and the TUAB studies of Appendices~\ref{app:prior-sensitivity} and \ref{app:mechanism} use width $64$ ($14.9$M) and $12$k steps.
The noise scale is $\sigma{=}0.02$ ($0.008$ on TUEV, see Appendix~\ref{app:sigma-sweep}), the prior uses $\tau{=}1$, $\alpha{=}2$, and $k{=}4$, the Hungarian coupling of SF2M and OT-CFM is solved exactly on each batch, and RF trains two rounds of the stated step count with one reflow on $8$k generated pairs between them.

\paragraph{Metrics.}
Metrics compare $512$ generated windows with $512$ held-out test windows at $200$\,Hz ($128$ for MEG, iEEG, and PEMS-BAY, at the sampling rates and splits of Appendix~\ref{app:ieeg-per-patient}).
In the eight-dataset sweep, each model is sampled across three generation seeds; reported values average over the five training seeds and three generation sets.
For PSD-KL, log band power is computed via Welch's method~\citep{welch1967psd} with Hann segments of $\min(256, T)$ samples and $50\%$ overlap ($200$ samples on SEED-V).
Because the logarithm is the transform that brings absolute band power closest to normal~\citep{gasser1982logtransform}, and for resting alpha it makes the power of single windows near-normal across windows~\citep{smulders2018logalpha}, we fit a univariate Gaussian to each channel and band, which gives the symmetric KL divergence between the real and generated fits in closed form, $\frac{1}{2}[D_{\mathrm{KL}}(\mathrm{real}\parallel\mathrm{gen}) + D_{\mathrm{KL}}(\mathrm{gen}\parallel\mathrm{real})]$, averaged across channels and bands.
For wPLI, cross-channel phase synchrony is evaluated via $\mathrm{wPLI}=|\mathbb{E}[\mathrm{Im}\,S_{xy}]|/\mathbb{E}[|\mathrm{Im}\,S_{xy}|]$~\citep{vinck2011wpli} using Welch cross-spectra ($512$-sample segments).
Using only the imaginary component of coherency~\citep{nolte2004imcoh}, wPLI eliminates spurious coupling from instantaneous volume conduction.
We correlate off-diagonal entries between real and generated matrices and report the band-averaged Pearson correlation (see Table~\ref{tab:wpli-main}); Gaussian white noise yields correlations within $\pm 0.03$ across all eight datasets.
We do not report the Fr\'echet Inception Distance because multi-channel EEG lacks a standardized feature extractor across diverse montages, and prior work found that the model with the most realistic spatial and spectral properties received the worst FID~\citep{hartmann2018eeggan}.

\paragraph{Downstream classifier.}

The EEGNet-8,2 classifier of Section~\ref{sec:downstream} follows \citet{lawhern2018eegnet}, namely $500$ epochs with validation-loss early stopping, MaxNorm $1$ on the depthwise layer and $0.25$ on the head, temporal kernel length $f_s/2$, Adam at learning rate $0.001$, dropout $0.25$, and five seeds.
The generator is the sweep-configuration U-Net with a class embedding at its input, trained with classifier-free guidance~\citep{ho2022cfg} (embedding dropped with probability $0.1$) and sampled with the embedding set to the target class.
To verify that the generator does not memorize training samples, we evaluate nearest-neighbor distances.
After standardizing each window, the $L_2$ distance from a generated window to its nearest real training window has median $183$ for either source, the same as between two real training windows ($180$), and its minimum ($121$ and $131$ for the isotropic and graph sources) is far above the real-to-real minimum ($19$).
The classifier is trained on real windows augmented with one generated window per training sample. Because raw generated amplitudes differ across priors, each synthetic window is scaled to match the average root-mean-square amplitude of the real training windows before concatenation.

\section{A Temporal Mat\'ern Source Without Spatial Structure}
\label{app:temporal-source}
The temporal source draws $x_0 = z\,\mathbf{V}^\top$ with $z\in\R^{C\times T}$ and independent entries $z_{ck}\sim\N(0,\psi(\omega_k))$, where $\mathbf{V}$ is the DCT-II basis of the window, $\omega_k=k/T$, and $\psi(\omega)=(1+\tau_t\omega)^{-\alpha_t}$ with $\tau_t{=}1$, $\alpha_t{=}2$, normalized to unit mean.
This prior imposes temporal smoothness while leaving spatial channels uncorrelated.
Combining it with the graph factor gives a separable spatiotemporal source whose covariance is the Kronecker product of the two, with samples $x_0=\Ubf\,\mathrm{diag}(\sqrt{\phi})\,z\,\mathrm{diag}(\sqrt{\psi})\,\mathbf{V}^\top$ for a standard normal matrix $z\in\R^{C\times T}$.
With SI at the sweep configuration on FACED, this source achieves a PSD-KL of $27.80 \pm 8.00$ across five seeds, compared to $29.03$ for the graph source alone (see Table~\ref{tab:sweep-all-main}).
Because this difference falls within cross-seed variance and the $\gamma$ band still accounts for $52\%$ of this source's error (compared with $55\%$ for GP-SI at $\sigma{=}0.10$; see Appendix~\ref{app:sweep-all}), temporal smoothing alone does not resolve the $\gamma$-band discrepancy.

Table~\ref{tab:temporal-source} reports PSD-KL at the sweep configuration (channel width $48$, $6$k steps, five training seeds, test split) for SI, SF2M, OT-CFM, and RF on the seven sweep datasets.
TUAB is omitted because its row in Table~\ref{tab:sweep-all-main} uses the larger model configuration.
The isotropic and graph baselines for SI and SF2M were retrained for this experiment and match the main benchmark within seed variance across most datasets.
On TUEV, the retrained isotropic baselines reach a higher PSD-KL than in Table~\ref{tab:sweep-all-main} ($3.81$ vs.\ $2.55$ for SI, and $4.32$ vs.\ $2.03$ for SF2M) despite identical noise scaling ($\sigma{=}0.008$).
An independent repetition yields SI $4.10 \pm 1.07$ and SF2M $3.45 \pm 0.62$, consistent with the training sensitivity documented in Appendix~\ref{app:failure-modes}.
The relative ranking among the three priors remains identical across runs.
Baseline entries for OT-CFM and RF are transcribed directly from the main sweep.
RF temporal models follow the standard two-stage reflow protocol ($6$k steps per round, $8$k reflow pairs) with the temporal source drawn in both stages.
Evaluations of wPLI show that the temporal source leaves the wPLI correlation at or below the isotropic level.
For SI on SEED-V, the band-averaged correlation is $-0.01$ for both temporal and isotropic sources, compared to $0.06$ for the graph prior.
On PhysioNet-MI, the temporal source yields $0.26$, compared to $0.32$ for isotropic noise and $0.44$ for the graph prior.

\begin{table}[t!]
\centering
\caption{PSD-KL (lower is better; mean$\pm$std over five training seeds) for the isotropic, graph-Mat\'ern, and temporal-Mat\'ern sources at the sweep configuration. Bold marks the best source within each method. The OT-CFM and RF isotropic and graph values are those of Table~\ref{tab:sweep-all-main}.}
\label{tab:temporal-source}
\vspace{0.1cm}
\setlength{\tabcolsep}{3pt}
\adjustbox{max width=\linewidth}{%
\begin{tabular}{l | ccc |ccc |ccc |ccc}
\toprule
& \multicolumn{3}{c|}{SI} & \multicolumn{3}{c|}{SF2M} & \multicolumn{3}{c|}{OT-CFM} & \multicolumn{3}{c}{RF} \\
\cmidrule(lr){2-4}\cmidrule(lr){5-7}\cmidrule(lr){8-10}\cmidrule(lr){11-13}
dataset (ch) & isotropic & graph & temporal & isotropic & graph & temporal & isotropic & graph & temporal & isotropic & graph & temporal \\
\midrule
TUEV (16)        & $3.81\!\pm\!1.25$ & $3.92\!\pm\!1.29$ & $\mathbf{2.69\!\pm\!0.92}$ & $4.32\!\pm\!1.34$ & $5.28\!\pm\!2.99$ & $\mathbf{4.24\!\pm\!1.19}$ & $5.30\!\pm\!2.30$ & $\mathbf{3.80\!\pm\!0.59}$ & $4.04\!\pm\!1.16$ & $3.19\!\pm\!1.21$ & $\mathbf{3.09\!\pm\!0.29}$ & $3.35\!\pm\!1.40$ \\
Mumtaz (19)      & $3.55\!\pm\!2.09$ & $3.58\!\pm\!1.05$ & $\mathbf{2.27\!\pm\!0.61}$ & $4.68\!\pm\!2.92$ & $4.14\!\pm\!1.90$ & $\mathbf{2.22\!\pm\!0.93}$ & $4.97\!\pm\!2.06$ & $5.51\!\pm\!1.66$ & $\mathbf{1.99\!\pm\!0.48}$ & $6.07\!\pm\!3.02$ & $3.79\!\pm\!1.33$ & $\mathbf{3.18\!\pm\!1.80}$ \\
BCI-IV 2a (22)   & $\mathbf{3.39\!\pm\!0.62}$ & $3.58\!\pm\!0.53$ & $3.69\!\pm\!0.38$ & $3.74\!\pm\!0.88$ & $3.57\!\pm\!1.29$ & $\mathbf{2.91\!\pm\!0.48}$ & $\mathbf{3.06\!\pm\!0.73}$ & $3.61\!\pm\!0.19$ & $3.30\!\pm\!0.50$ & $4.27\!\pm\!0.71$ & $\mathbf{4.24\!\pm\!1.35}$ & $4.76\!\pm\!0.64$ \\
FACED (32)       & $33.34\!\pm\!2.87$ & $\mathbf{28.81\!\pm\!4.55}$ & $32.86\!\pm\!5.93$ & $32.83\!\pm\!3.49$ & $31.84\!\pm\!7.15$ & $\mathbf{31.34\!\pm\!2.00}$ & $33.60\!\pm\!7.20$ & $31.50\!\pm\!12.00$ & $\mathbf{29.68\!\pm\!6.00}$ & $32.14\!\pm\!3.91$ & $\mathbf{26.44\!\pm\!4.96}$ & $29.30\!\pm\!13.33$ \\
SHU (32)         & $113.46\!\pm\!2.43$ & $\mathbf{83.17\!\pm\!3.71}$ & $125.63\!\pm\!3.17$ & $112.68\!\pm\!3.26$ & $\mathbf{83.42\!\pm\!4.21}$ & $125.83\!\pm\!2.50$ & $111.27\!\pm\!5.19$ & $\mathbf{85.36\!\pm\!0.71}$ & $124.99\!\pm\!3.67$ & $115.72\!\pm\!2.26$ & $\mathbf{82.88\!\pm\!10.33}$ & $125.63\!\pm\!1.77$ \\
SEED-V (62)      & $32.94\!\pm\!0.73$ & $\mathbf{27.99\!\pm\!0.15}$ & $36.80\!\pm\!0.80$ & $32.85\!\pm\!0.75$ & $\mathbf{28.02\!\pm\!0.21}$ & $36.75\!\pm\!0.92$ & $32.90\!\pm\!0.70$ & $\mathbf{28.10\!\pm\!0.30}$ & $36.74\!\pm\!0.82$ & $33.33\!\pm\!0.38$ & $\mathbf{28.26\!\pm\!0.21}$ & $36.68\!\pm\!0.80$ \\
PhysioNet-MI (64) & $32.94\!\pm\!0.87$ & $\mathbf{19.57\!\pm\!2.26}$ & $36.13\!\pm\!1.86$ & $34.21\!\pm\!0.75$ & $\mathbf{20.44\!\pm\!3.25}$ & $38.01\!\pm\!1.46$ & $33.00\!\pm\!1.40$ & $\mathbf{19.90\!\pm\!3.90}$ & $38.27\!\pm\!0.70$ & $33.34\!\pm\!1.83$ & $\mathbf{22.44\!\pm\!1.54}$ & $38.38\!\pm\!2.64$ \\
\bottomrule
\end{tabular}}
\end{table}

\section{Participation Ratio vs Tikhonov Effective Dimension on TUAB}
\label{app:pr-vs-tikhonov}

The main text uses the participation-ratio (PR) effective dimension $d_{\mathrm{eff}}(\Sbf_0) = (\tr\Sbf_0)^2/\tr(\Sbf_0^2)$, which has no free parameter, to measure how concentrated the prior's spectrum is.
On TUAB it falls from $16$ for the isotropic prior to $10.1$ for the graph-Mat\'ern prior.
The quantity that enters a linear-Gaussian variance bound for flow matching is instead the Tikhonov effective dimension $d_{\mathrm{eff},\lambda}(\Sbf_0) = \tr(\Sbf_0(\Sbf_0+\lambda\Id)^{-1})$, where $\lambda$ is a regularization strength.
Between the two priors it differs by far less at every $\lambda$, from $15.84$ against $15.78$ at $\lambda{=}0.01$ to $8.00$ against $7.19$ at $\lambda{=}1$ (see Table~\ref{tab:pr-vs-tikhonov}), so the PR value overstates how much the prior lowers the variance term.

\begin{table}[ht]
\centering
\caption{PR and Tikhonov effective dimensions of the isotropic and graph-Mat\'ern priors on TUAB, both normalized to $\tr\Sbf_0=C$.
The graph-Mat\'ern value is smaller at every $\lambda$, but by far less than in the PR.}
  \vspace{0.1cm}
\label{tab:pr-vs-tikhonov}
\adjustbox{max width=\linewidth}{%
\begin{tabular}{lrr}
\toprule
quantity & isotropic & graph-Mat\'ern\\
\midrule
PR $d_{\mathrm{eff}}$ & $16.00$ & $10.09$ \\
\midrule
Tikhonov $\lambda{=}0.01$ & $15.84$ & $15.78$\\
Tikhonov $\lambda{=}0.10$ & $14.55$ & $14.08$ \\
Tikhonov $\lambda{=}0.50$ & $10.67$ & $9.75$ \\
Tikhonov $\lambda{=}1.00$ & $8.00$  & $7.19$ \\
Tikhonov $\lambda{=}2.00$ & $5.33$  & $4.82$ \\
\bottomrule
\end{tabular}}
\end{table}

\section{Graph-Prior Hyperparameter Sensitivity}
\label{app:prior-sensitivity}
Table~\ref{tab:sensitivity} reports a $3\times 3$ sweep of $(\tau, \alpha)$ at fixed $k{=}4$ on TUAB with SI.
Every setting except $(4, 4)$ lies in $[1.13, 1.81]$, with the best, $(0.25, 4)$, at $1.13 \pm 0.08$ and the default $(1, 2)$ at $1.39 \pm 0.05$.
The sweep is a separate set of five training seeds at channel width $64$ and $12$k steps on the validation split with a cosine learning-rate schedule, which is why its default entry differs from the graph-Mat\'ern references of Table~\ref{tab:prior-ablation} ($1.08 \pm 0.41$) and Appendix~\ref{app:precond-baselines} ($1.54 \pm 0.91$), both trained with a constant learning rate on their own seeds; the three values are within one another's seed variance.
The configuration $(\tau{=}4, \alpha{=}4)$ degrades performance ($18.10 \pm 4.80$) because it puts almost all of the source variance on the lowest mode, so the source is nearly rank one and the drift does not learn to transport it to the data within the training budget.
We keep $(1, 2)$ on every dataset rather than the best setting for two reasons.
It is a generic setting of the graph-Mat\'ern density of \citet{borovitskiy2021graph}, namely decay rate $\alpha{=}2$ (the Mat\'ern smoothness) and correlation length $\tau{=}1$, the unit scale of the normalized Laplacian's eigenvalue range $[0, 2]$, fixed once rather than fitted by marginal likelihood as in that work.
This generic setting demonstrates that the prior provides consistent gains without per-dataset tuning.
In addition, the failure at $(\tau{=}4, \alpha{=}4)$ shows that $\alpha{=}4$, which the best setting shares, leaves no safety margin for unseen sensor layouts.

\begin{table}[h]
\centering
\caption{Validation PSD-KL on TUAB with SI for the $3{\times}3$ $(\tau, \alpha)$ sweep at $k{=}4$ (mean$\pm$std). Default $(1, 2)$ in bold.}
\vspace{0.1cm}
\label{tab:sensitivity}
\adjustbox{max width=\linewidth}{%
\begin{tabular}{lccc}
\toprule
$\tau$ $\backslash$ $\alpha$ & $1$ & $2$ & $4$ \\
\midrule
$0.25$ & $1.62 \pm 0.33$ & $1.36 \pm 0.42$ & $1.13 \pm 0.08$ \\
$1.0$  & $1.46 \pm 0.37$ & $\mathbf{1.39 \pm 0.05}$ & $1.57 \pm 0.15$ \\
$4.0$  & $1.44 \pm 0.31$ & $1.81 \pm 0.22$ & $18.10 \pm 4.80$ \\
\bottomrule
\end{tabular}}
\end{table}

\section{Noise-Scale (\texorpdfstring{$\sigma$}{sigma}) Calibration Across Datasets}
\label{app:sigma-sweep}

We calibrate the noise scale through the ratio $\kappa{=}\sigma/\sigma_{\mathrm{real}}$, where $\sigma_{\mathrm{real}}$ is the per-channel standard deviation on the training split.
Setting $\sigma{=}0.08\,\sigma_{\mathrm{real}}$ reduces per-dataset tuning to a single signal-scale estimate.
A sweep of $\kappa$ from $0.02$ to $0.32$ on TUAB with SI at the configuration of Appendix~\ref{app:mechanism} (channel width $64$, $12$k steps, validation split, five seeds) finds no value between $0.04$ and $0.32$ distinguishable from the default $0.08$ for either source, as cross-seed variance is comparable to the mean differences; only the smallest value, $\kappa{=}0.02$, raises the isotropic mean to $2.50$.
The calibration rule is therefore robust to at least a factor-of-two variation around $0.08$, defining the stable interval $[0.04, 0.16]$.

Table~\ref{tab:kappa-check} lists the $\sigma$ used on each dataset together with its $\sigma_{\mathrm{real}}$ and the implied $\kappa$.
In practice, $\sigma{=}0.02$ was used on every dataset except TUEV.
FACED, whose $\sigma_{\mathrm{real}}$ is comparable to TUEV's, used $\sigma{=}0.02$ ($\kappa{=}0.194$), placing it above this interval.
All other datasets fall within the interval $\kappa \in [0.04, 0.16]$, clustering closely around the target ratio of $0.08$.
TUEV's rescaled $\sigma{=}0.008$ gives $\kappa{=}0.082$, matching TUAB's $\kappa{=}0.081$ at $\sigma{=}0.02$.

\begin{table}[t!]
\centering
\caption{$\kappa = \sigma/\sigma_{\mathrm{real}}$ across datasets at the $\sigma$ used in the eight-dataset sweep.
FACED's $\kappa{=}0.194$ lies outside the plateau $[0.04, 0.16]$ on which the rule was validated.
Its per-band results are in Appendix~\ref{app:sweep-all}.
}
\vspace{0.1cm}
\label{tab:kappa-check}
\adjustbox{max width=\linewidth}{%
\begin{tabular}{lccr}
\toprule
dataset & $\sigma$ & $\sigma_{\mathrm{real}}$ & $\kappa$ \\
\midrule
TUAB         & 0.02  & 0.247 & 0.081 \\
TUEV         & 0.008 & 0.098 & 0.082 \\
Mumtaz-MDD   & 0.02  & 0.215 & 0.093 \\
BCI-IV 2a    & 0.02  & 0.336 & 0.060 \\
FACED        & 0.02  & 0.103 & 0.194 \\
SHU          & 0.02  & 0.271 & 0.074 \\
SEED-V       & 0.02  & 0.246 & 0.081 \\
PhysioNet-MI & 0.02  & 0.280 & 0.071 \\
\bottomrule
\end{tabular}}
\end{table}

\section{Graph Prior Ablation Study}
\label{app:mechanism}
Unless noted otherwise, every experiment in this section trains SI on TUAB at the configuration of its row in Table~\ref{tab:sweep-all-main} (dim $64$, $12$k steps) and reports PSD-KL on the validation split.
The isotropic ($1.85 \pm 0.18$) and graph-Mat\'ern ($1.08 \pm 0.41$) reference values therefore differ from the test-split values in that table ($1.43$ and $1.16$).
We use SI because its independent coupling adds no alignment step between the source and the drift that could mask a change of the source.

\subsection{Replacing Only the Eigenvectors: Shuffled Positions and Random Bases}
\label{app:random-eigenbasis}
Rotating the 3-D sensor coordinates does not change the graph prior, since rigid rotations preserve pairwise Euclidean distances, the $k$-NN graph, and its eigenvectors $U$ in channel space.
To isolate the eigenvectors from the spectrum, the three priors in this group retain the Mat\'ern density $\phi(\lambda)=(1+\tau\lambda)^{-\alpha}$ and normalization unchanged ($d_{\mathrm{eff}}{=}10.1$), altering only the spatial directions.

\paragraph{Shuffled sensor positions.}
Permuting sensor coordinates before constructing the $k$-NN graph attaches the same graph topology and spectrum to incorrect channels.
Across $25$ random permutations, the mean PSD-KL is $3.70 \pm 2.79$ (median $2.71$).
$64\%$ of the permutations perform worse than the uninformative isotropic source ($1.85 \pm 0.18$), $84\%$ are worse than the true position graph ($1.08 \pm 0.41$), and the spread across permutations is nearly seven times larger ($2.79$ vs.\ $0.41$).
The wide spread plausibly reflects how much local neighborhood structure each permutation happens to preserve.

\paragraph{Uniformly random orthogonal basis.}
Replacing $U$ with a uniformly random orthogonal matrix $Q \in \mathbb{R}^{C\times C}$ preserves the exact Mat\'ern eigenvalues but randomizes channel axes, which gives $4.42 \pm 2.97$, worse than the isotropic source.

\paragraph{Empirical covariance eigenvectors.}
Pairing the eigenvectors of the empirical data covariance $\hat\Sigma_{\mathrm{emp}}$ with the smooth graph-Mat\'ern spectrum yields $2.76 \pm 2.93$.
Even with a smoothly decaying spectrum and no small eigenvalues, data-driven PCA modes fail to substitute for the physical sensor eigenbasis.

\subsection{Concentrating Variance in the Wrong Directions: Empirical Covariance and Shrinkage}
\label{app:learned-prior}
\label{app:lw-prior}
The empirical-covariance prior tests whether a concentrated source spectrum suffices without montage geometry.
We estimate $\hat\Sigma_{\mathrm{emp}}$ over all time samples of the TUAB training windows, eigendecompose it as $\hat\Sigma_{\mathrm{emp}} = U_{\mathrm{emp}}\Lambda_{\mathrm{emp}} U_{\mathrm{emp}}^\top$, scale $\Lambda_{\mathrm{emp}}$ so that its mean eigenvalue is one as for the graph prior, and draw $x_0 = U_{\mathrm{emp}}\,\mathrm{diag}(\sqrt{\lambda_{\mathrm{emp}}})\,z$.
This prior uses no montage geometry and is more concentrated than the graph-Mat\'ern prior ($d_{\mathrm{eff}}{=}5.2$ against $10.1$, with two eigenvalues below $0.001$).
If concentration alone explained the gain, this prior would outperform the graph-Mat\'ern prior.
Instead, its PSD-KL is $4.61 \pm 2.40$, worse than both the isotropic source ($1.85 \pm 0.18$) and all priors that replace only the eigenvectors (see Table~\ref{tab:prior-ablation}).
Concentrating variance is therefore not enough, since the variance has to follow the sensor layout.

\paragraph{Regularization via Ledoit--Wolf shrinkage.}
Because $\hat\Sigma_{\mathrm{emp}}$ is estimated from finite training data, we test whether ill-conditioned or noisy sample eigenvalues cause this failure.
We regularize the covariance as $\hat\Sigma_{\mathrm{LW}} = (1-\rho)\hat\Sigma_{\mathrm{emp}} + \rho\bar\lambda\,\Id$, where $\bar\lambda = \mathrm{tr}(\hat\Sigma_{\mathrm{emp}})/C$ and $\rho$ is the optimal Ledoit--Wolf shrinkage intensity~\citep{ledoit2004well}.
Shrinking the empirical covariance flattens its spectrum and raises the effective dimension from $5.2$ to $6.6$, but preserves its eigenvectors.
At $3.56 \pm 2.43$, the regularized prior remains worse than isotropic noise ($1.85 \pm 0.18$), confirming that the failure lies in the data-driven eigenvectors rather than eigenvalue noise.

\paragraph{Synthetic Gaussian target.}
To test whether covariance mismatch between a Gaussian source and the target explains the failure, we replace the EEG target with a synthetic Gaussian target, white in time, matching the empirical TUAB channel covariance.
On this target, all six tested Gaussian sources reach a similar PSD-KL ($0.0020$--$0.0032$; unit isotropic $0.0020 \pm 0.0004$, empirical covariance $0.0022 \pm 0.0005$, Ledoit--Wolf $0.0022 \pm 0.0004$).
The velocity field absorbs any mismatch between Gaussian distributions, so the empirical prior's failure on real EEG likely comes from temporal dependencies and non-Gaussian structure that this target lacks.

\subsection{Spectrum Shape: Smoothness and Mode Coverage}
\label{app:spectrum-shape}
We next evaluate the functional form of the spectral roll-off while holding the eigenvectors of the position graph fixed.

\paragraph{Heat-kernel spectrum.}
Replacing the polynomial Mat\'ern density with an exponential heat kernel $\phi(\lambda) = \exp(-\tau\lambda)$ on the same Laplacian yields $1.38 \pm 0.86$.
This retains most of the benefit of the graph-Mat\'ern prior ($1.08 \pm 0.41$), although placing more variance on higher-frequency modes ($d_{\mathrm{eff}}{=}12.0$ vs.\ $10.1$) raises the standard deviation over seeds from $0.41$ to $0.86$.
Appendix~\ref{app:prior-sensitivity} provides a sensitivity analysis over the Mat\'ern hyperparameters $(\tau, \alpha)$.

\paragraph{Hard low-pass spectrum.}
Assigning equal variance to the $C/2 = 8$ smoothest eigenmodes and setting the remaining upper half strictly to zero ($d_{\mathrm{eff}}{=}8.0$) fails ($71.40 \pm 12.50$).
Because flow matching learns a smooth invertible map, it cannot map a source with zero variance on a subspace to a target that has variance in every direction.

\subsection{Graph Sparsity and Construction: Fully Connected vs.\ Local Graphs}
\label{app:prior-structure}
We examine how graph connectivity and construction influence generation.

\paragraph{Fully connected Gaussian graph.}
Connecting every sensor pair with a dense Gaussian weight $\exp(-\lVert p_i-p_j\rVert^2/h^2)$ rather than restricting edges to $k$-NN yields $2.53 \pm 1.81$ ($d_{\mathrm{eff}}{=}11.2$), performing worse than isotropic noise.
Geometric distance alone is therefore insufficient; the inductive bias requires the eigenvectors of a sparse, local graph.

\paragraph{Correlation graph in place of coordinates.}
Replacing coordinate distances with training-set cross-channel correlations $|\mathrm{corr}(x_i, x_j)|$ in a $k{=}4$ nearest-neighbor graph yields $1.39 \pm 1.70$ across ten seeds (median $0.90$, $d_{\mathrm{eff}}{=}11.0$).
The correlation graph shares only $36\%$ of its edges with the spatial coordinate graph; its strongest edges link distant cross-scalp channels (\textit{e.g.}, frontal--occipital pairs reaching $|\mathrm{corr}| \approx 0.99$), likely due to broad-band artifacts shared across the montage rather than local coupling.
A graph built from sensor positions has no such long-range edges.

\subsection{Whitening and a Graph Convolution in the Drift}
\label{app:precond-baselines}
Two baselines place spatial structure in the data or in the drift instead of in the source.
The first whitens every window with a fixed zero-phase (ZCA) whitening map $\hat\Sigma^{-1/2}$~\citep{bell1997independent,kessy2018optimal}, the symmetric inverse square root of the channel covariance estimated from $1024$ training windows, and trains with the isotropic source on the whitened data, undoing the map before evaluation.
The second keeps the drift's 1D U-Net and inserts a channel mixer $(\Id-\beta\Laplacian)$ with a learned scalar $\beta$ at its input and output, with $\Laplacian$ the same $k$-NN Laplacian the prior uses, under either source.
All five configurations of Table~\ref{tab:precond-baselines} are trained with SI on TUAB in one run per seed, so they share the same training batches, evaluation set, and isotropic reference, with the bipolar-pair coordinates, the validation split, batch size $64$, and five seeds at the section's configuration (dim $64$, $12$k steps).
The five configurations are within the spread over seeds of one another, isotropic $1.23 \pm 0.36$, whitened $0.98 \pm 0.38$, graph-Mat\'ern $1.54 \pm 0.91$, and $1.46 \pm 0.54$ and $1.67 \pm 1.23$ with the mixer, so at this budget placing spatial structure in the data or in the drift does not separate from the isotropic reference on TUAB.
The isotropic reference of this run is itself below the section's ($1.85 \pm 0.18$), which comes from a separate set of training seeds.

\begin{table}[t!]
\centering
\caption{PSD-KL on the TUAB validation split (lower is better; mean$\pm$std over five training seeds) for the whitening and graph-mixer baselines with SI at dim $64$, $12$k steps.
Bold denotes the lowest mean; every difference is within the spread over seeds.}
\label{tab:precond-baselines}
\vspace{0.1cm}
\adjustbox{max width=\linewidth}{%
\begin{tabular}{lc}
\toprule
configuration & PSD-KL \\
\midrule
isotropic source, plain drift              & $1.23 \pm 0.36$ \\
ZCA-whitened data, isotropic source        & $\mathbf{0.98 \pm 0.38}$ \\
graph-Mat\'ern source, plain drift         & $1.54 \pm 0.91$ \\
isotropic source, graph-mixer drift        & $1.46 \pm 0.54$ \\
graph-Mat\'ern source, graph-mixer drift   & $1.67 \pm 1.23$ \\
\bottomrule
\end{tabular}}
\end{table}

\begin{table}[t!]
\centering
\caption{Per-patient PSD-KL on the CCEP ECoG dataset (mean$\pm$std over five training seeds; lower is better).
The channel counts, the number of ECoG channels with valid 3-D coordinates, are $104$ (UMCU01), $88$ (UMCU02), $72$ (UMCU03), $112$ (UMCU04), and $96$ (UMCU05).
Bold denotes the better source within each method for each patient and for the mean.}
\vspace{0.1cm}
\label{tab:ieeg-per-patient}
\setlength{\tabcolsep}{3pt}
\adjustbox{max width=\linewidth}{%
\begin{tabular}{l cc cc cc cc}
\toprule
& \multicolumn{2}{c}{$\sigma{>}0$ Hungarian} & \multicolumn{2}{c}{$\sigma{>}0$ independent} & \multicolumn{2}{c}{$\sigma{=}0$ Hungarian} & \multicolumn{2}{c}{$\sigma{=}0$ indep+reflow} \\
\cmidrule(lr){2-3}\cmidrule(lr){4-5}\cmidrule(lr){6-7}\cmidrule(lr){8-9}
patient & SF2M & GP-SF2M & SI & GP-SI & OT-CFM & GP-OT-CFM & RF & GP-RF \\
\midrule
UMCU01 & $58.04{\pm}0.64$ & $\mathbf{35.54{\pm}1.81}$ & $57.52{\pm}2.23$ & $\mathbf{35.91{\pm}1.74}$ & $56.24{\pm}1.51$ & $\mathbf{34.90{\pm}1.90}$ & $67.38{\pm}0.75$ & $\mathbf{48.55{\pm}0.89}$ \\
UMCU02 & $5.97{\pm}0.17$ & $\mathbf{3.40{\pm}0.10}$ & $5.89{\pm}0.29$ & $\mathbf{3.62{\pm}0.32}$ & $5.77{\pm}0.17$ & $\mathbf{3.60{\pm}0.09}$ & $9.04{\pm}0.90$ & $\mathbf{5.01{\pm}0.40}$ \\
UMCU03 & $2.47{\pm}0.12$ & $\mathbf{1.57{\pm}0.05}$ & $2.49{\pm}0.09$ & $\mathbf{1.71{\pm}0.08}$ & $2.45{\pm}0.07$ & $\mathbf{1.55{\pm}0.09}$ & $3.71{\pm}0.20$ & $\mathbf{2.31{\pm}0.08}$ \\
UMCU04 & $76.39{\pm}1.13$ & $\mathbf{47.62{\pm}1.62}$ & $75.75{\pm}2.02$ & $\mathbf{44.79{\pm}2.46}$ & $74.89{\pm}1.69$ & $\mathbf{47.27{\pm}3.14}$ & $89.92{\pm}1.41$ & $\mathbf{58.24{\pm}3.54}$ \\
UMCU05 & $13.46{\pm}0.32$ & $\mathbf{7.78{\pm}0.18}$ & $14.18{\pm}0.89$ & $\mathbf{7.45{\pm}0.26}$ & $13.00{\pm}0.69$ & $\mathbf{7.28{\pm}0.43}$ & $17.99{\pm}0.46$ & $\mathbf{10.49{\pm}0.40}$ \\
\midrule
\textbf{mean across patients} & $31.27$ & $\mathbf{19.18}$ & $31.17$ & $\mathbf{18.69}$ & $30.47$ & $\mathbf{18.92}$ & $37.61$ & $\mathbf{24.92}$ \\
\bottomrule
\end{tabular}}
\end{table}

\section{Details of the Other Modalities}
\label{app:ieeg-per-patient}
The three modalities of Section~\ref{sec:ieeg} use the protocol of Section~\ref{subsec:protocol} with the graph and the training setup of each modality given below.
Table~\ref{tab:ieeg-per-patient} gives the per-patient values behind the iEEG row of Table~\ref{tab:cross-modality}.

\paragraph{Whole-head MEG.}
We use the Wakeman--Henson face-perception cohort~\citep{wakemanhenson2015}, $16$ subjects recorded with an Elekta Vectorview helmet ($102$ magnetometers, one graph shared across subjects), resampled to $250$\,Hz and segmented into $4$\,s windows.
One model per method, source, and seed is trained on all $16$ subjects for $48$k steps and evaluated on the last $20\%$ of each subject's recording.
Unlike the EEG runs, SF2M and SI on MEG are sampled with $\sigma\sqrt{\Delta t}$ Gaussian noise added after each Euler step.
All three modalities train a three-level U-Net of channel width $48$ (channel multipliers $1$, $2$, $4$) with batch size $32$, $\sigma{=}0.02$, and a cosine learning-rate schedule, and evaluate $128$ generated against $128$ held-out windows.
The step count is chosen where PSD-KL improvements stabilize, as extending training from $12$k to $48$k steps shifts relative gains by at most $0.3$ percentage points across methods.

\paragraph{Intracranial EEG.}
We use five CCEP ECoG~\citep{vanblooijs2023ccep} patients (UMCU01--05) whose ECoG grids have $72$ to $112$ channels (see Table~\ref{tab:ieeg-per-patient}).
Each patient has a unique grid and channel count, so a shared Laplacian is not defined.
We train one model per patient, method, source, and seed on that patient's recordings (band-pass $0.5$--$100$\,Hz, $50$\,Hz notch, $250$\,Hz sampling rate, 8\,s windows) using the graph Laplacian constructed from that patient's electrode coordinates.
Each model trains for $6$k steps; RF trains two rounds of $6$k steps with one reflow on $2{,}048$ generated pairs.

\paragraph{PEMS-BAY traffic.}
The sensor network of \citet{li2018dcrnn} has $325$ freeway loop sensors in the San Francisco Bay Area.
We convert latitude and longitude to Earth-centered coordinates for the $k$-NN graph and use windows of $T{=}256$ samples.
PEMS-BAY is sampled every $5$ minutes, so its Nyquist frequency of $1.67$\,mHz lies far below the EEG bands, and PSD-KL instead uses five uniform bands of width $0.33$\,mHz.
One model per method, source, and seed trains for $6$k steps; RF trains two rounds of $6$k steps with one reflow on $4{,}096$ generated pairs.
The symmetrized $k{=}4$ graph is connected (its Laplacian has a single zero eigenvalue) on the $325$-sensor network and on every ECoG grid, which we verify for each layout before training.

\begin{figure}[t!]
\centering
\includegraphics[width=0.85\linewidth]{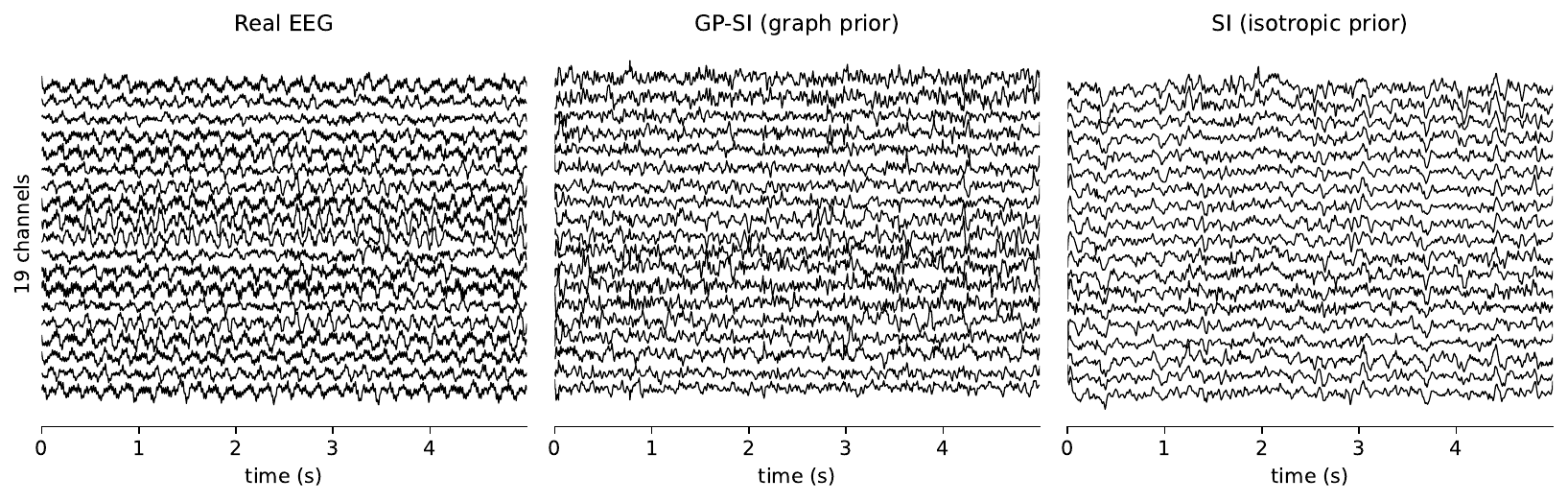}
\caption{\textbf{Qualitative effect of the graph prior on Mumtaz-MDD} (19 channels). Real EEG (left), GP-SI (middle), and SI (right). GP-SI reproduces the slow-wave dynamics and channel coupling of real signals, whereas isotropic SI produces decoupled channel noise.}
\label{fig:qual_mumtaz}
\end{figure}

\begin{figure}[t!]
\centering
\includegraphics[width=0.85\linewidth]{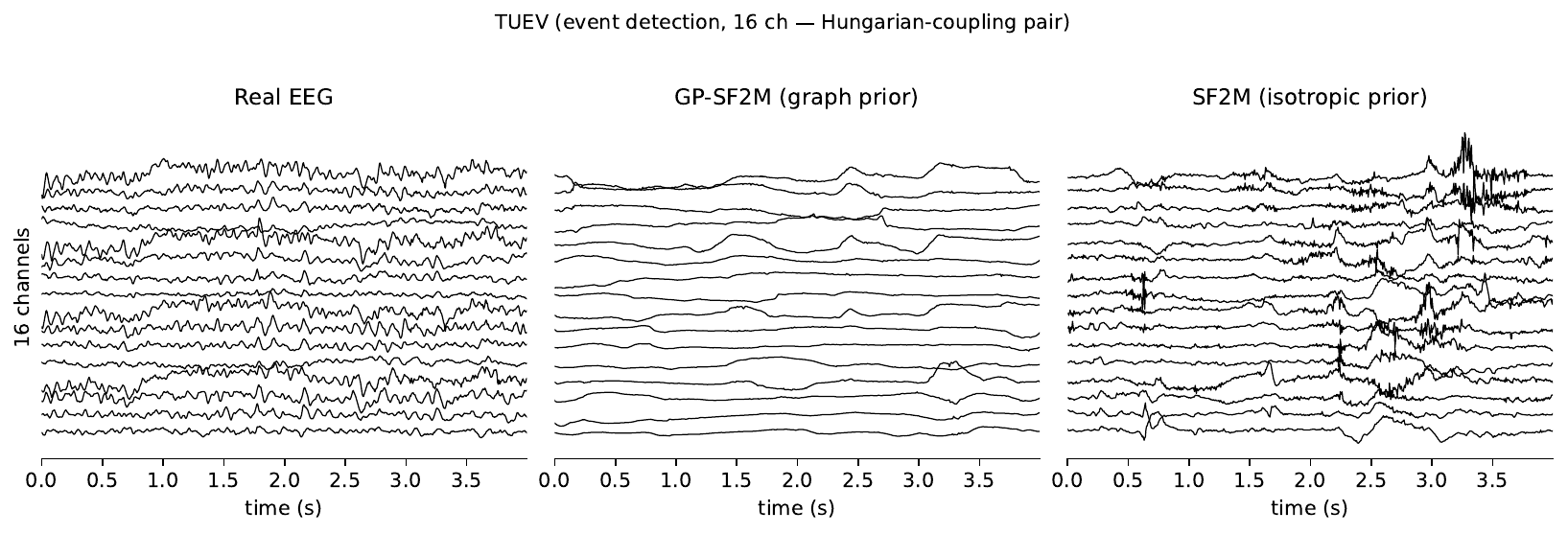}
\caption{\textbf{Qualitative effect of the graph prior on TUEV} (16 channels). Real EEG (left), GP-SI (middle), and SI (right). GP-SI reproduces the slow-wave dynamics and channel coupling of real signals, whereas isotropic SI produces decoupled channel noise.}
\label{fig:qual_tuev}
\end{figure}

\section{Qualitative and Per-Band Details of the Eight-Dataset Sweep}
\label{app:sweep-all}
The eight-dataset sweep of Table~\ref{tab:sweep-all-main} trains the four methods with the isotropic and the graph source on each dataset.

\paragraph{Per-dataset qualitative waveforms.}
Figures~\ref{fig:qual_mumtaz} and \ref{fig:qual_tuev} repeat the qualitative comparison of Figure~\ref{fig:qual_tuab} on Mumtaz and TUEV.
On Mumtaz (depression, 19 channels) the graph prior retains the slow waves and inter-channel coherence of real EEG, whereas the isotropic prior produces noisier channels that are only weakly coupled.
On TUEV, the graph prior preserves the slow-wave amplitude variability of real EEG and inter-channel coupling, while the isotropic prior produces unstructured noise with decoupled channels.

\paragraph{Per-band breakdown on SHU and SEED-V.}
Table~\ref{tab:perband-multi} reports per-band PSD-KL on SHU (32 channels) and SEED-V (62 channels) for all four methods, with the $\sigma{>}0$ methods at the default $\sigma{=}0.02$.
The SF2M, SI, and OT-CFM rows are computed from the per-seed band values of the benchmark runs in Table~\ref{tab:sweep-all-main} (five training seeds, three generated sets each), so their avg column matches that table.
The RF rows come from a separate five-seed run of the same reflow protocol whose per-band values were stored; their avg column ($114.53$ and $82.02$ on SHU, $33.30$ and $28.50$ on SEED-V) sits within the cross-seed variance of the RF and GP-RF values in Table~\ref{tab:sweep-all-main} ($115.72 \pm 2.26$ and $82.88 \pm 10.33$; $33.33 \pm 0.38$ and $28.26 \pm 0.21$).
Every method with the isotropic source yields comparable averages regardless of coupling ($32.84$ to $33.30$ on SEED-V and $111.27$ to $114.53$ on SHU), and every method with the graph source yields consistently lower error ($28.01$ to $28.50$ on SEED-V and $81.36$ to $85.36$ on SHU).
The source prior matters more than the choice of coupling, since the four graph-source averages lie within $2\%$ of each other on SEED-V and within $5\%$ on SHU, while the gap between the two sources is $14\%$ to $15\%$ on SEED-V and $23\%$ to $28\%$ on SHU.
On SHU, the graph source reduces PSD-KL across all frequency bands for every method, with the $\gamma$ term dropping from $340$ to $261$ for SI.

\begin{table}[t!]
\centering
\caption{Per-band PSD-KL on SHU and SEED-V for all four methods (test split; the $\sigma{>}0$ methods at the default $\sigma{=}0.02$); bold marks the best value per band per dataset.}
\label{tab:perband-multi}
\setlength{\tabcolsep}{4pt}
\adjustbox{max width=0.9\linewidth}{%
\begin{tabular}{lllrrrrrr}
\toprule
dataset & variant & coupling & $\delta$ & $\theta$ & $\alpha$ & $\beta$ & $\gamma$ & avg \\
\midrule
\multirow{8}{*}{SEED-V (62ch)} & SF2M     & Hung, $\sigma{>}0$    & 1.27 & 7.75 & 14.18 & 56.20 & 84.82 & 32.84 \\
& GP-SF2M  & Hung, $\sigma{>}0$    & 1.25 & 6.41 & 11.66 & 47.20 & 73.73 & 28.05 \\
& SI       & indep, $\sigma{>}0$   & 1.31 & 7.78 & 14.20 & 56.36 & 85.01 & 32.93 \\
& GP-SI    & indep, $\sigma{>}0$   & \textbf{1.22} & \textbf{6.40} & \textbf{11.65} & \textbf{47.14} & \textbf{73.66} & \textbf{28.01} \\
& OT-CFM   & Hung, $\sigma{=}0$    & 1.28 & 7.79 & 14.22 & 56.29 & 84.90 & 32.90 \\
& GP-OT-CFM& Hung, $\sigma{=}0$    & 1.26 & 6.41 & 11.67 & 47.22 & 73.76 & 28.06 \\
& RF       & indep+reflow, $\sigma{=}0$ & 1.32 & 7.93 & 14.47 & 57.14 & 85.63 & 33.30 \\
& GP-RF    & indep+reflow, $\sigma{=}0$ & 1.26 & 6.53 & 11.85 & 47.76 & 75.08 & 28.50 \\
\midrule
\multirow{8}{*}{SHU (32ch)} & SF2M     & Hung, $\sigma{>}0$    & 6.04 & 24.38 & 28.87 & 168.65 & 337.45 & 113.08 \\
& GP-SF2M  & Hung, $\sigma{>}0$    & 3.57 & 16.67 & 18.71 & \textbf{111.25} & 256.59 & \textbf{81.36} \\
& SI       & indep, $\sigma{>}0$   & 6.10 & 24.59 & 29.01 & 171.71 & 340.40 & 114.36 \\
& GP-SI    & indep, $\sigma{>}0$   & \textbf{3.31} & \textbf{16.27} & \textbf{18.39} & 114.40 & 261.43 & 82.76 \\
& OT-CFM   & Hung, $\sigma{=}0$    & 5.83 & 23.81 & 28.04 & 162.59 & 336.05 & 111.27 \\
& GP-OT-CFM& Hung, $\sigma{=}0$    & 3.41 & 16.53 & 18.99 & 122.27 & 265.61 & 85.36 \\
& RF       & indep+reflow, $\sigma{=}0$ & 6.47 & 25.13 & 29.76 & 173.86 & 337.45 & 114.53 \\
& GP-RF    & indep+reflow, $\sigma{=}0$ & 3.61 & 16.43 & 19.31 & 116.05 & \textbf{254.71} & 82.02 \\
\bottomrule
\end{tabular}}
\end{table}

\begin{figure}[t!]
\centering
\vspace{-0.2cm}
\includegraphics[width=0.7\linewidth]{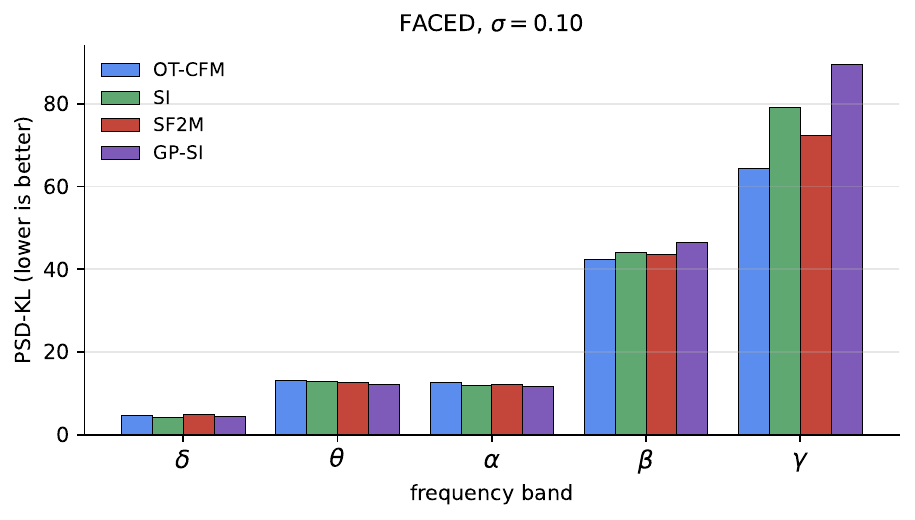}
\vspace{-0.2cm}
\caption{Per-band PSD-KL on FACED for OT-CFM ($\sigma{=}0$) and for SI, SF2M, and GP-SI at $\sigma{=}0.10$ (mean over seeds; lower is better).}
\label{fig:faced-perband}
\end{figure}

\paragraph{Per-band breakdown on TUEV and FACED.}
\label{app:failure-modes}
TUEV is the only dataset on which the graph prior makes SF2M worse ($2.03 \to 3.85$ at the sweep configuration; see Table~\ref{tab:sweep-all-main}).
The increase is spread over all five bands (from $+82\%$ in $\delta$ and $\alpha$ to $+241\%$ in $\beta$, five seeds at $\sigma{=}0.008$), and the cross-seed variance of GP-SF2M is comparable to the shift, indicating training variance rather than a localized spectral failure.
On FACED the prior lowers PSD-KL for all four methods at $\sigma{=}0.02$ (see Table~\ref{tab:sweep-all-main}).
At $\sigma{=}0.10$, where high noise exacerbates spectral divergence, GP-SI matches or outperforms isotropic baselines in $\delta$, $\theta$, and $\alpha$, but diverges in $\beta$ and $\gamma$ (see Figure~\ref{fig:faced-perband}); $\gamma$ alone contributes more than half of the five-band average.
A separable spatiotemporal source on FACED is reported in Appendix~\ref{app:temporal-source}.

\end{document}